\documentclass{article}
\usepackage{arxiv}
\usepackage[utf8]{inputenc}  
\usepackage[T1]{fontenc}     
\usepackage{amsmath,amsfonts}
\usepackage[english]{babel}
\usepackage[xspace]{ellipsis}
\usepackage{graphicx}
\usepackage{enumitem}
\usepackage{booktabs}

\usepackage[hidelinks]{hyperref}        
\usepackage{url}             
\usepackage{float}           
\usepackage{comment}
\usepackage{nicefrac}        
\usepackage{microtype}       
\usepackage[title]{appendix} 
\usepackage{multirow}        
\usepackage{lipsum}
\usepackage{todonotes}      
\usepackage{setspace}       
\usepackage{color, colortbl} 
\usepackage{soul}
\usepackage{subcaption}
\usepackage{natbib}

\graphicspath{ {./figures/}}
\usepackage{defs}

\title{Abolitionist Networks: Modeling Language Change in Nineteenth-Century Activist Newspapers}

\author{Sandeep Soni \\
	School of Interactive Computing\\
	Georgia Institute of Technology\\
	\texttt{sandeepsoni@gatech.edu} \\
	\And
	Lauren Klein\thanks{Also affiliated to the Department of English at Emory University} \\
	Department of Quantitative Theory and Methods\\
	Emory University\\
	\texttt{lauren.klein@emory.edu} \\
	\AND
	Jacob Eisenstein \\
	Google Research \\
	\texttt{jeisenstein@google.com} \\
}
\date{}
\renewcommand{\undertitle}{Published in the Journal of Cultural Analytics}
\renewcommand{\shorttitle}{Soni, Klein, and Eisenstein / JCA 2021}

\begin{document}


\maketitle


\begin{abstract}
The abolitionist movement of the nineteenth-century United States remains among the most significant social and political movements in US history. Abolitionist newspapers played a crucial role in spreading information and shaping public opinion around a range of issues relating to the abolition of slavery. These newspapers also serve as a primary source of information about the movement for scholars today, resulting in powerful new accounts of the movement and its leaders. This paper supplements recent qualitative work on the role of women in abolition’s vanguard, as well as the role of the Black press, with a quantitative text modeling approach. 
Using diachronic word embeddings, we identify which newspapers tended to lead lexical semantic innovations --- the introduction of new usages of specific words --- and which newspapers tended to follow. We then aggregate the evidence across hundreds of changes into a weighted network with the newspapers as nodes; directed edge weights represent the frequency with which each newspaper led the other in the adoption of a lexical semantic change. Analysis of this network reveals pathways of lexical semantic influence, distinguishing leaders from followers, as well as others who stood apart from the semantic changes that swept through this period. More specifically, we find that two newspapers edited by women --- \newspaper{The Provincial Freeman} and \newspaper{The Lily} --- led a large number of semantic changes in our corpus, lending additional credence to the argument that a multiracial coalition of women led the abolitionist movement in terms of both thought and action. It also contributes additional complexity to the scholarship that has sought to tease apart the relation of the abolitionist movement to the women’s suffrage movement, and the vexed racial politics that characterized their relation.
\end{abstract}

\section{Introduction}
\label{sec:introduction}

Scholars often grapple with the power they hold as the producers of historical knowledge, as well as with the historical forces of power that shaped the documents that provide them with evidence about the past~\citep[e.g.,][]{Trouillot_1995,Stoler_2009,Fuentes_2016}. These forces of power are plainly apparent in the newspapers that document the abolitionist movement of the nineteenth-century United States. These newspapers played a crucial role in spreading information and shaping public opinion about the abolition of slavery and related social justice issues. They also now serve as a primary source of information about abolition for scholars today. And yet, these newspapers are still subject to the historical forces that surrounded their production: those edited by white people (and white men in particular) have been more fully preserved, and therefore, are more accessible to researchers. As a result, until relatively recently, these newspapers --- and the efforts of the white men who edited them --- have dominated historical accounts of the abolitionist movement \citep{Ernest_2004}. However, in the past decade or so, scholars~\citep{Foster_2005,Gardner_2011,Fagan_2016,Spires_2019} 
have worked to highlight the leadership role of the Black press in the movement. This has resulted in new and necessary historical narratives that confirm the central role played by the Black press in the fight against slavery and in support of liberation broadly conceived.   

\begin{figure}
    \centering
    \begin{subfigure}{0.55\textwidth}
    \includegraphics[width=.9\textwidth]{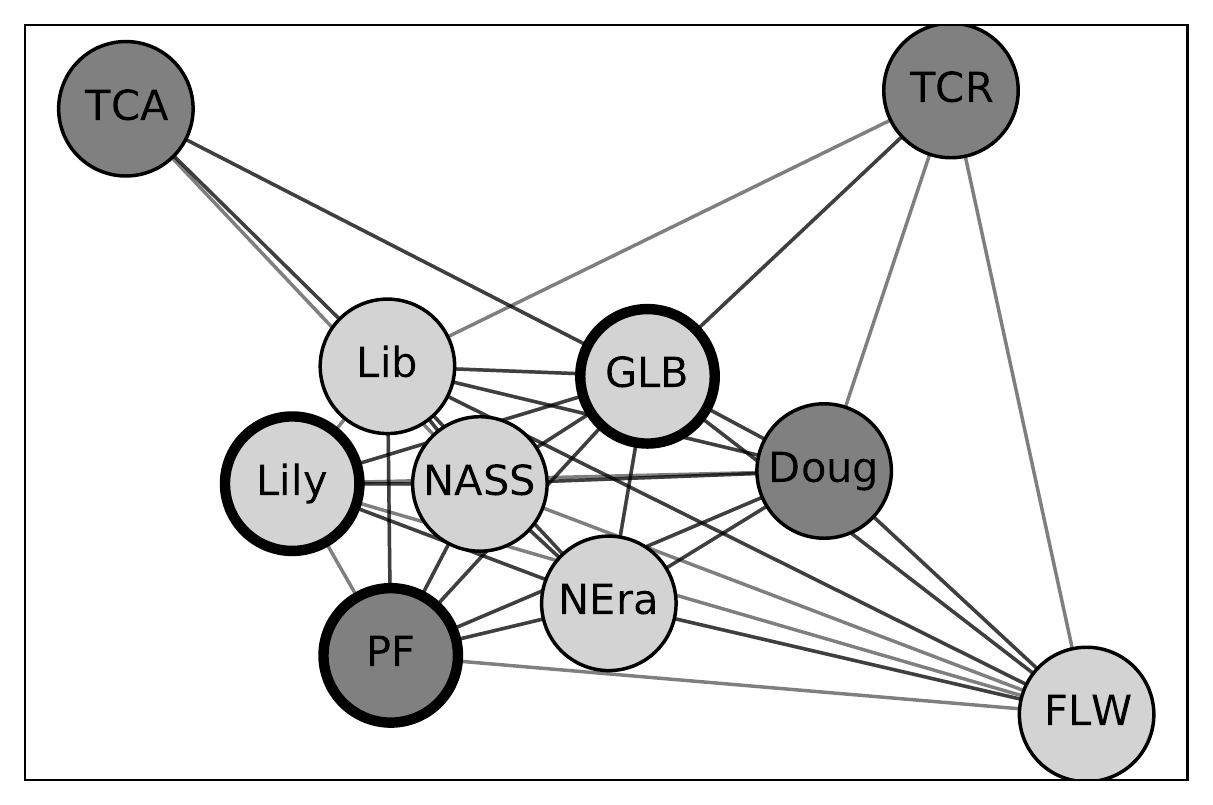}
    \end{subfigure}
    \begin{subtable}{0.4\textwidth}
    \small
    \begin{tabular}{ll}
         \textsf{Doug} & \newspaper{Douglass Newspapers}\\
         \textsf{FLW} & \newspaper{Frank Leslie's Weekly}\\
         \textsf{GLB} & \newspaper{Godey's Lady's Book}\\
         \textsf{Lib} & \newspaper{The Liberator}\\
         \textsf{Lily} & \newspaper{The Lily}\\
         \textsf{NASS} & \newspaper{National Anti-Slavery Standard}\\
         \textsf{NEra} & \newspaper{The National Era}\\
         \textsf{PF} & \newspaper{The Provincial Freeman}\\
         \textsf{TCA} & \newspaper{The Colored American}\\
         \textsf{TCR} & \newspaper{The Christian Recorder}
    \end{tabular}
    \end{subtable}
    
    \caption{
    A spring layout of the ten newspapers in our collection, in which newspapers are positioned nearby in space to the extent that they influence each other;\citep{fruchterman1991graph}
    the strength of the pairwise influence is also reflected in the darkness of the tie. Dark gray nodes represent members of the Black press; nodes with darker outlines represent publications edited by and/or aimed at women. For more details on the newspapers, see \nameref{sec:data}.
    }
    \label{fig:spring_layout}
\end{figure}

But there is more work to be done. Intersectional approaches to the study of abolition have highlighted the role of women, and Black women in particular, in the movement's vanguard. Citing newspapers edited by women, both Black and white, among other sources, Manisha Sinha, for example, makes the case that ``women were abolition's foot-soldiers and, more controversially, its leaders and orators''~\citep[][pg. 2]{sinha2016slave}. This work dovetails with the past several decades of scholarship on nineteenth-century Black women's activism, which has demonstrated how Black women emerged as ``visible and authoritative community leaders'' over the course of the nineteenth century~\citep[][pg. 2]{Jones_2009}. Yet it has been generally understood that this emergence was gradual, with Black women securing their public presence in the years after the Civil War \citep[e.g.,][]{Peterson_1995,Foreman_2009,Jones_2009}. While recovery efforts such as the Colored Conventions Project~\citeyear{About_2020} 
have sought to excavate evidence of how Black women participated in early conversations about abolition and related issues of social, economic, and educational justice, it is still often necessary to ``read between the lines'' of the archival record ``for echoes of women's contributions'' to the abolitionist movement, as Derrick Spires explains~\citep[][pg. 108]{Spires_2019}.

This project shares the commitment of the sources cited above to amplifying the leadership contributions of Black people, and Black women in particular, to the abolitionist cause. But we diverge in our methodological approach. Rather than seek out additional archival sources that might augment existing accounts of social and political leadership --- in terms of either thought, or action, or both --- or ``read between the lines'' of existing documents so as to generate new narratives about who exemplified leadership in its various forms, we focus on the \emph{micro-structure} of the language of the archive. We hypothesize that aspects of social and political leadership --- more specifically, the introduction of new concepts, the reframing of existing ones, and the advancement and circulation of both --- can be identified in the appearance of individual words and the statistics that describe their lexical contexts. By  connecting these words and contexts to the specific newspapers in which they appear, we offer a new layer of evidence about the emergence of certain social and political currents, the sources responsible for their transmission, and the networks that helped sustain them.\footnote{The code and some analysed data can be found at \url{https://github.com/sandeepsoni/semantic-leadership-network}. The supporting data for the analysis can be found at \url{https://doi.org/10.7910/DVN/EWYMFG}} 

Contextual accounts of the meaning of words, such as the one just described, are known in linguistics as \term{distributional semantics}~\citep{firth1957synopsis}. Through computational analysis of large text corpora, this contextual information can be summarized in the form of numerical vectors known as \textbf{word embeddings}, which have been shown to provide a powerful measure of changes in the meaning of individual terms~\citep{wijaya2011understanding,kutuzov2018diachronic}, as well as an index of deeper social dynamics~\citep{garg2018word,kozlowski2019geometry}. For these reasons, word embeddings have attracted growing interest in the digital humanities, where they have been applied to identify the evolution of concepts~\citep{Gavin_Jennings_Kersey_Pasanek_2019} and offer a ``lexical panorama'' of a particular discourse~\citep{Shechtman_2020}. We go further by adding a network perspective to the dynamics of word meaning: we ask not only \emph{which} individual words have changed in meaning and \emph{how} they have changed, but also \emph{who} was responsible for those changed meanings and how that responsibility was expressed. We propose a method of identifying the specific newspapers that were responsible for introducing new or changed meanings of words into the surrounding discourse, as well as the newspapers which quickly adopted those new or changed usages. To do so, we develop a model of \emph{semantic change} and a measure of \emph{semantic leadership} that links each word in the corpus to a pair of newspapers: a leader and a follower. We then aggregate over hundreds of semantic changes to construct a network between the set of newspapers in our collection. The analysis of this network reveals the specific roles played by each newspaper in the evolving discourse of abolitionism. This novel computational approach offers a counterweight, however imperfect, to the forces of power that operate along the lines of both race and gender, and that have overdetermined much of the historical scholarship on the abolitionist press to this point.

We find that two newspapers edited by women --- \newspaper{The Provincial Freeman} 
and \newspaper{The Lily} --- played a leading role in a large number of semantic changes in our corpus. That the former is an abolitionist newspaper edited by a Black woman, and the latter is a women's suffrage newspaper edited by a white woman, lends additional credence to the argument that a multiracial coalition of women led the abolitionist movement in terms of both thought and action. In addition, it contributes additional complexity to the scholarship that has sought to tease apart the relation of the abolitionist movement to the women's suffrage movement, and the vexed racial politics that characterize their relation~\citep[e.g.,][]{Jones_2020, McRae_2018}. By aggregating this evidence of semantic leadership into a weighted network, with the newspapers as nodes and the number of their lead words as edge weights, we reveal additional pathways of semantic influence. More specifically, we find evidence to confirm extant accounts of the leadership role played by \newspaper{The Liberator}, a white-edited newspaper, as well as new insight into the leadership relationships among newspapers associated with the Black press. Our methods of validation, which involve both qualitative and quantitative analysis, demonstrate how this evidence might be incorporated back into existing scholarly conversations about nineteenth-century freedom struggles. More generally, our results suggest that our model of semantic leadership might be applied to other incomplete or otherwise fragmentary archives, thereby contributing to the rebalancing of the historical narratives that the archives prompt.

\section{Lexical Semantic Change and Leadership}
\label{sec:methodoverview}

Our methodological approach draws on both semantic and sociolinguistic perspectives on language change,
which have rarely been reconciled in prior work.
This may be due in part to each subfield's preferred sources of data: semantic change is studied mainly from diachronic corpora of written texts, which typically have limited social information, while sociolinguistic research is built mainly on interviews, with temporal changes inferred from each speaker's age~\citep{bailey2002real}. Because our newspaper corpus (described more fully in \nameref{sec:data}) is large and diachronic, and contains sources that were known to have been in conversation, we are able to take a step towards unifying these perspectives in a novel modeling framework (described in \nameref{sec:method}). To set the stage for this work, we now review the key foundations of our modeling framework before summarizing our methodological contributions.

\subsection{Lexical semantic change}
\label{sec:related-lexsem}
Languages undergo constant change, and even an intuitive reflection on a word like \example{awesome} makes clear that this instability extends to the lexicon, which links words and their semantics~\citep{pierrehumbert2012dynamic}. Lexical semantic changes can be identified in corpora by building on the theory of distributional semantics~\citep{firth1957synopsis}, which asserts that the meaning of a linguistic element can be ascertained by the contexts in which it appears. Consequently, if a word undergoes a change in the distribution over contexts --- in its distributional statistics --- then its meaning is likely to have changed as well. In contemporary computational linguistics, distributional statistics are summarized by vector word embeddings, which are usually estimated by maximizing a probability model on a corpus~\citep{mikolov2013distributed}. In the past several years, word embeddings have begun to be applied in digital humanities research~\citep[e.g.][]{schmidt2015vector, heuser2017word, Gavin_2018, Gavin_Jennings_Kersey_Pasanek_2019}.
By making word embeddings dynamic, they can be powerful tools for revealing changes in word meaning~\citep[e.g.,][]{wijaya2011understanding,kim2014temporal,kulkarni2015statistically,hamilton2016diachronic,rudolph2018dynamic,yao2018dynamic}. A typical approach is to estimate embeddings on multiple corpora from different time periods and then align the embedding vectors to make them comparable; however, there are many alternatives~\citep[for an overview, see][]{kutuzov2018diachronic, tahmasebi2018survey,tang2018state}. 

After identifying lexical semantic changes, we next ask what can be learned from them in aggregate. \cite{hamilton2016diachronic} treat diachronic word embeddings as evidence for structural constraints on semantic change, while \cite{garg2018word} trace shifts in public attitudes towards race and ethnicity by tracking word embeddings in large-scale corpora of books. In the digital humanities, diachronic word embeddings have been used to trace the history of concepts: for example, \cite{Shechtman_2020} explores the ``technical, ideological, and environmental valences'' of the idea of media by comparing near neighbors of \example{media} and related terms across a corpus of 20th century magazines. Most closely related to our own work, \cite{soni2020follow} present a method for identifying the specific documents that lead lexical semantic changes, which the authors show to accrue more citations in corpora of scientific abstracts and legal opinions.
This work provides further evidence for the validity of our approach of linking lexical semantic leadership to broader influence. Methodologically, we diverge from this prior work by identifying leaders of individual semantic changes, and then computing and analyzing aggregate leadership networks over hundreds of changes.


\subsection{The leaders of language change and their social networks}
\label{sec:related-socio}
The search for leaders of language change is a core concern of sociolinguistics, which seeks to identify groups of people who tend to be in the linguistic vanguard~\cite{labov2001principles}. A robust finding from this literature is that women tend to lead language change, particularly in phonology (the sound system) and in grammar~\citep[See][]{labov1990intersection,eckert2013language,tagliamonte2009peaks}. This finding is generally reported without consideration of non-binary genders; for research on the sociolinguistics of non-binary genders, see \cite{zimman2009language} and \cite{gratton2016resisting}.
This type of sociolinguistic research relies heavily on the method of \emph{apparent time}~\citep{bailey2002real}: first, identify changes in progress by comparing the speech of younger and older individuals; then, identify individuals whose speech is better aligned with that of younger people, which implies that they were among the earlier adopters of the change. The key assumption is that each person's use of language is stable throughout their adult life, a hypothesis that has been broadly supported by prior work on changes in phonology and grammar~\citep{sankoff2007language}. Here, however, we focus on changes in word meaning, where there is less evidence in favor of adult stability.\footnote{\cite{danescu2013no} find evidence of stability in \emph{which} words are used, but provide no evidence about word meaning.} 
We therefore provide \emph{real-time} evidence from a diachronic corpus, and we identify many changes in the lexical semantics of individual editors.

Methodologically, we analyze language change in the context of a social network. Prior work in sociolinguistics has linked the network characteristics of individuals with their positioning with respect to language change: for example, \cite{labov2001principles} argues that individuals with many weak ties tend to be leaders, while \cite{milroy1987language} finds that dense sub-networks of strong ties tend to resist language change. Our focus is not on characterizing the social networks of the leaders of language change, but on aggregating dyadic leader-follower relationships into an overall summary of the pathways of innovation in a community of newspapers. More closely related is computational research on lexical innovations in Twitter: \cite{eisenstein2014diffusion} construct a network of sociolinguistic influence between cities in the United States, and \cite{goel2016social} model the structural characteristics of influence events between individual Twitter users. Both of these studies are based on the \emph{frequencies} of lexical items, rather than changes in their meanings. 

Beyond sociolinguistics, there have been a number of attempts to quantify influence in text corpora, largely from researchers in computer science and machine learning. Some approaches operate at the level of individual word frequencies: for example, \cite{guo2015bayesian} measure influence within small groups by the use of individual words, on the assumption that Alice is likely to have influenced Bob if words used by Alice tend to be subsequently used by Bob.
\cite{kelly2018measuring} apply this idea to a dataset of patents, finding that the most influential patents (in terms of citations and market value) are those whose word distributions are distinct from prior documents, but similar to subsequent documents. 
Topic models~\cite{blei2012probabilistic} have also been used to measure influence in networks~\citep{tang2009social,liu2010mining}: in a humanities context, \cite{barron2018individuals} apply a topic-based analysis to speeches from the French revolution, identifying as influencers those individuals whose topic distributions were distinct from the past and similar to the future. 
All of these methods focus on changes in which words are used, rather than how they are used. While lexical frequencies can be informative --- particularly in cases that fit theoretical models like communicative accomodation theory~\citep{giles1991accommodation} --- they are often caused by outside events (e.g., new people or keywords) rather than conceptual changes in the discourse.

To summarize, prior work in sociolinguistics has been concerned largely with phonology and morphosyntax, while computational researchers have focused on the frequency of words or topics. In contrast, we study changes in the meanings of words, as quantified by diachronic word embeddings that summarize the changing contexts in which those words appear. 
Such changes are especially relevant to questions of social and political influence in a networked discourse --- as in the various groups involved in the abolitionist movement of the nineteenth-century United States. For example, it has long been argued that Black Americans, both those enslaved and those free, helped to expand the idea of what \example{freedom} properly entails \citep[e.g.][]{Kelley_2002,Tillet_2012}. While freedom, in the nineteenth century, was initially understood in terms of legal emancipation --- a conception formulated by white abolitionists fixated on ending slavery --- Black Americans of that same era understood freedom as much more capacious, encompassing ideas about social, economic, and educational justice, as well as liberation broadly conceived. Is it possible to detect a signal that tracks this conceptual change? And if so, is it possible to determine who in the corpus was responsible for advancing and/or popularizing the expanded conception of the term? Our method allows us to explore these questions from a semantic and statistical perspective: we can test whether words like \example{freedom} really did undergo semantic change in the documents in our newspaper corpus; identify the specific newspaper that led those changes; and aggregate across individual changes to an overall summary of lexical semantic leadership.
This aggregate measure functions as a proxy for overall linguistic influence in the corpus, taking into account the influence of known words and phrases, like \example{freedom}, as well as other terms that might not otherwise invite attention in a manual analysis. The resulting network provides a new way of understanding the relationships among the newspapers that document this pivotal era in the history of the United States. 

\subsection{Summary of technical contributions}
The work described in this paper builds upon these efforts to model semantic changes and aggregate them into a summary network. We enhance prior diachronic word embedding techniques by incorporating metadata about the newspapers that serve as the source of word occurrence. We then compute the lead or lag of each source with respect to each semantic change. These \emph{lead-lag pairs} are then aggregated into a network of semantic leadership, which we analyze to arrive at a new understanding of the roles of each newspaper in the linguistic changes that accompanies the evolving discourse on the abolition of slavery in the United States. To summarize, we make three methodological contributions: 
\begin{enumerate}
\item We propose a text modeling approach to identify semantic leadership in a corpus of timestamped documents from a set of sources, such as newspapers. As described in the following section, this approach includes (a) a model to identify semantic changes in language using diachronic word embeddings, and (b) a statistical measure to quantify the extent to which each source led others in the adoption of each change.
\item We apply this method to a corpus of nineteenth-century newspapers digitized by Accessible Archives~\citep{maret2016accessible}. We extract the words that changed in meaning during the period between 1827 to 1865, the period bounded by the publication of the first known abolitionist newspaper and the ratification of the Emancipation Proclamation.
We then quantify the lead-lag relationships between the newspapers for each of these changes.
\item We aggregate the dyadic relationships between newspapers into a weighted semantic leadership network, retaining links between newspapers for each semantic change only if the relationship is so strong as to be highly unlikely to have arisen by chance. We then apply two metrics for holistic network analysis --- PageRank~\citep{page1999pagerank} and Hyperlink Induced Topic Search (HITS)~\citep{kleinberg1999authoritative} --- to identify the specific roles of each newspaper as a leader, follower, or outsider in the evolving discourse of abolitionism.
\end{enumerate}

\section{Method: From timestamped text to aggregate influence networks}
\label{sec:method}
This section describes our method in more detail. We first describe a model to learn diachronic word embeddings from text(see \nameref{sec:method-temporal-change})
, then augment the model to learn diachronic word embeddings that are conditional on the newspapers and which can be used to statistically measure the lead of one newspaper over other (see \nameref{sec:method-source-conditional})
, and finally propose a scheme to aggregate the lead-lag relationships of newspapers and quantify their influence(see \nameref{sec:method-control-randomness} and \nameref{sec:method-aggregation-and-centrality})

\paragraph{Mathematical notation.} 

We define a document as a sequence of discrete tokens (words) from a finite vocabulary $\Vcal$, so that document $i$ is denoted $w_i = (w_{i,1}, w_{i,2}, \ldots, w_{i, n_i})$, with $n_i$ indicating the length of document $i$. We define a corpus as a set of $N$ documents, $\corp = \{w_1, w_2, \ldots, w_N\}$. Each document is associated with a discrete time $t_i$ and source label $s_i$. The time is created by binning the document timestamps into $T$ bins; there are $S$ sources, corresponding to each of the newspapers in the collection. The specific newspapers and time bins are described in the following section.

\subsection{Word embeddings}
Our approach builds on word embeddings, which represent each word in the vocabulary as a vector of real numbers~\citep{deerwester1990indexing,turney2010frequency}. We estimate the embeddings by optimizing the classical \term{skipgram} objective, in which embeddings are parameters in a model of the probability of each word token conditioned on its neighboring tokens~\citep{mikolov2013distributed}. While the method is well known, the details are relevant to the innovative aspects of our methodology, so we review them briefly. Omitting the document index $i$, the skipgram objective is based on the probability,
\begin{equation}
P(w_{j'} \mid w_j) \propto \exp \left( \embout_{w_{j'}} \cdot \embin_{w_j} \right),
\label{eq:skipgram-prob}
\end{equation}
where $\embout_{w_{j'}}$ is the ``output" embedding of $w_{j'}$, 
and $\embin_{w_j}$ is the ``input'' embedding of $w_{j}$. 
Each $\embout$ and $\embin$ is a vector, which is estimated by maximizing the product of these probabilities across all nearby pairs of tokens $(w_j, w_{j'})$. The result is that words that appear in similar contexts tend to have similar embedding vectors, which is the essence of the theory of distributional semantics~\citep{harris1954distributional,firth1957synopsis,lenci2008distributional}. We focus on the input embeddings $\embin$, as in most of the prior applications of the skipgram embedding model to digital humanities research \citep[e.g.,][]{heuser2017word,Gavin_2018}.

\subsection{Identifying semantic changes}
\label{sec:method-temporal-change}
The skipgram embedding model is not equipped to use metadata such as times associated with the tokens. However, \citep{bamman2014distributed} proposed an extension to specialize the embeddings accounting for any discrete labels attached to each token. The core idea of their approach is to decompose the input embedding of a word as a sum of two embeddings --- a base embedding and a residual embedding. Though originally developed to learn distinct embeddings of words based on geography, we apply it to learn diachronic embeddings~\citep[See also recent work by][]{gillani2019simple}. To do this, we discretize time into $T$ bins, with $t_i \in 1 \dots T$ indicating the time bin for document $i$. Then if $w$ appears at position $j$ in document $i$, the following input embedding is substituted in~\autoref{eq:skipgram-prob}.
\begin{equation}
\label{eq:temporal-residual}
\embin_{w_{i,j}}^{(t_i)} = \basein_{w_{i,j}} + \resin_{w_{i,j}}^{(t_i)},
\end{equation}
where $\basein_{w_{i,j}}$ is the base embedding of the word $w_{i,j}$ and $\resin_{w_{i,j}}^{(t_i)}$ is the residual for time $t_i$. The base embedding represents the elements of meaning that do not change over time; the residual embeddings are specialized for the contexts encountered at each temporal epoch, making it possible to model semantic change. To ensure that the base embedding is used to account for the time-invariant components of meaning, the residual embeddings are regularized by adding an $l_2$ penalty to the skipgram objective, corresponding to the sum of squared values, $\sum_{w}^{\Vcal} \sum_{t}^{T} (\resin_{w}^{(t)})^2$. This encourages the estimator to choose values of $\resin$ that are close to zero unless a strong temporal signal is observed.



\begin{figure}
    \centering
    \includegraphics[scale=0.6]{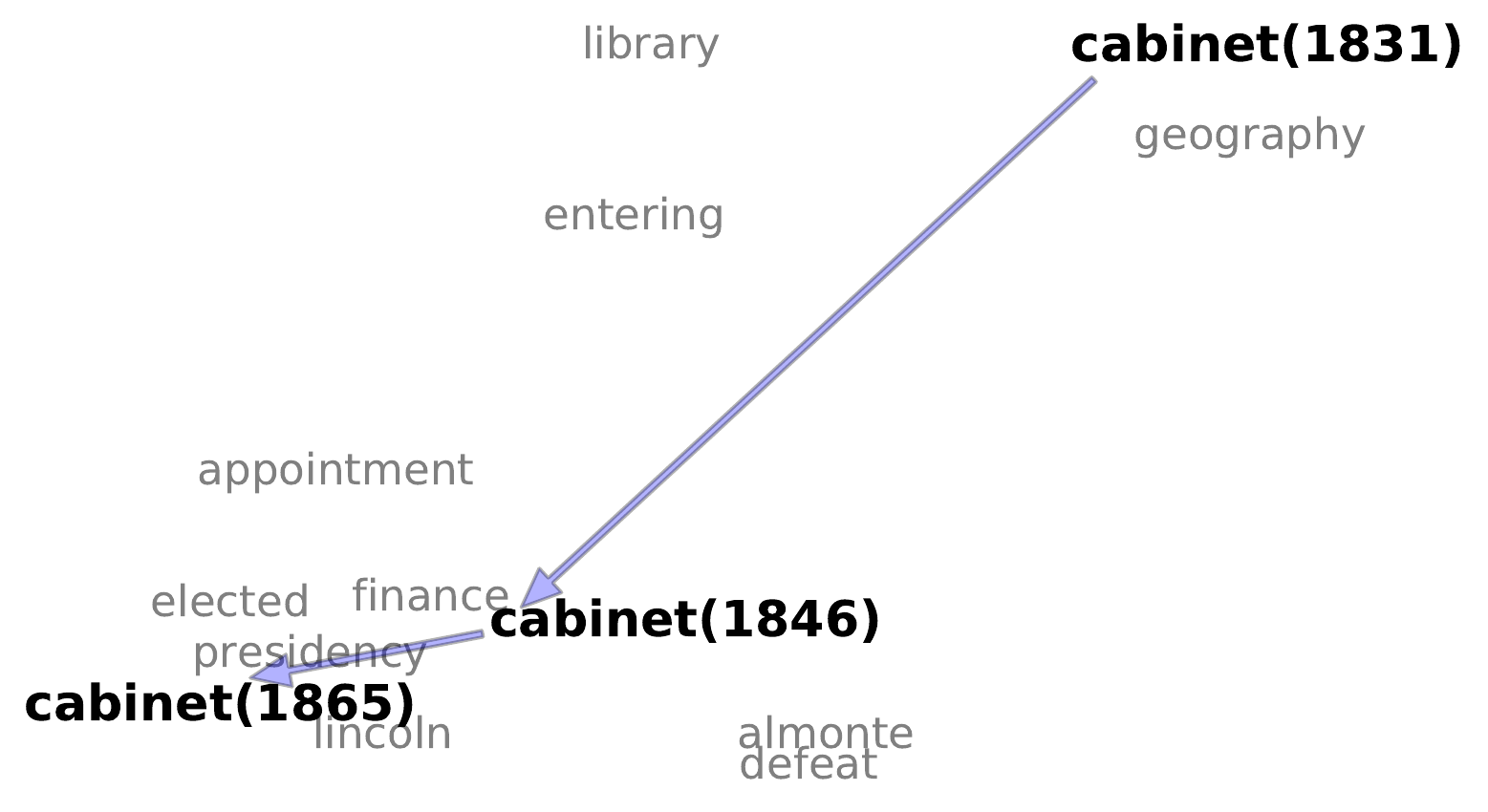}
    \caption{\textbf{Semantic change in \example{cabinet}.} The embeddings of \example{cabinet} and its near neighbors are projected to two dimensions using singular value decomposition. The embeddings of the target word indicate meaning shift from the architectural uses to political uses. This is reflected in the position of the embedding but even more so in the change in near neighbors.
    }
    \label{fig:semantic-change-example}
\end{figure}

We focus on terms whose embeddings change substantially over time. Among many possible ways to quantify change in embeddings~\citep[e.g.][]{hamilton2016cultural,asgari2020unsupervised,gonen2020simple}, we apply the method from~\citep{hamilton2016cultural}, in which the words are ranked for semantic change by computing the difference in the similarity of their neighbors over time in the embedding space. Because we have multiple time bins, we extend this method by computing the differences between neighbors for all pairs of times in $T$ for every word and pick the interval that has the maximum difference as the measure of semantic change in the word. Words are then ranked based on this measure, yielding an ordered list of changes as tuples of words and timestamps. 

\autoref{fig:semantic-change-example} provides an example: the term \example{cabinet} enters the corpus being used in the (now) archaic sense to describe a small storage room, as in a cabinet of curiosities; but in the final years of the corpus, fully shifts to reflect its usage in the political sense, as in the context of a president's Cabinet (other changes pertinent to abolition are described below in the section \nameref{sec:results-semantic-changes}). As a measure of the validity of our approach, many changes refer to entities and concepts from the Civil War, such as \example{ferry} (later referring to \example{Harper's Ferry}), \example{confederates} (acquiring a specialized meaning after the formation of the Confederacy), and \example{battery} (from an assault to an array of guns). Changes referring to specific named entities were filtered at a later stage.

Our approach of modeling semantic changes has three main advantages. First, as a joint model of words and time, it does not require the computationally expensive post-alignment of the word embeddings that is common in standard approaches~\citep[e.g.][]{kulkarni2015statistically,hamilton2016diachronic}. Second, temporal embeddings can be learned even for words that emerge or disappear before the start or end of the time period respectively --- these are just the base embeddings. Finally, the model is easily extended to incorporate other metadata about the text if available, as described in the next section.   

\subsection{Finding semantic leaders}
\label{sec:method-source-conditional}
Language changes are not adopted by everyone at the same time: for any change there are leaders, followers, and individuals who resist the change altogether. Moreover, being a leader for one semantic change does not imply semantic leadership for all changes~\citep{raumolin2006leaders}. We are interested in identifying who led and followed in each semantic change. To do this, we augment the diachronic embedding model to include an additional term for the \emph{source} of each token --- that is, the newspaper in which the token appears. The underlying input embedding representation can then be written,
\begin{equation}
\label{eq:temporal-residual}
\embin_{w_{i,j}}^{(t_i,s_i)} = \basein_{w_{i,j}} + \resin_{w_{i,j}}^{(t_i)} + \resin_{w_{i,j}}^{(t_i,s_i)},
\end{equation}
where $\resin_{w_{i,j}}^{(t_i,s_i)}$ is the source-specific temporal deviation added to the temporal and atemporal components of the input embedding. As before we apply $l_2$ regularization to the residual terms, which ensures that $\resin_{w}^{(t, s)} = 0$ for words that are not used by source $s$ at time $t$.

The fully conditioned embeddings are then used to assign a leadership score between a pair of sources for a given change $(w, t_1, t_2)$, where $t_1 < t_2$ are the timestamps of a change in the meaning of word $w$. For each pair of sources, $s_1$ and $s_2$, we calculate two quantities: first, the similarity in meaning of the word for $s_1$ with $s_2$ at the two different times and, second, the similarity in meaning of the word for $s_2$ at the two different times. A measure of the lead of source $s_1$ over $s_2$ is the ratio of vector inner products,
\begin{equation}
\label{eq:lead-equation}
    \lead(\edge{s_1}{s_2},w,t_1,t_2) = \frac{\embin_{w}^{(t_1,s_1)} \cdot \embin_{w}^{(t_2, s_2)}}{\embin_{w}^{(t_1,s_2)} \cdot \embin_{w}^{(t_2,s_2)}}
\end{equation}
A higher score indicates more leadership, with a score of $1$ corresponding to a baseline case of no leadership. For more intuition, consider a few cases:
\begin{itemize}
    \item If $s_1$ is indeed leading $s_2$ on $w$, then $s_1$'s use of the word with the new meaning should precede that of $s_2$. \autoref{eq:lead-equation} checks this precisely by comparing cross-source similarity (the numerator) with self-similarity (the denominator) in meaning across the two times. If the numerator is larger than the denominator, then $s_1$  is likely to have used the word with the new meaning before $s_2$ adopted it. Conversely, if the denominator is larger, then $s_2$ has not followed $s_1$ in its use of this word.
    \item If $w$ does not change in meaning, then no newspaper should be deemed a leader.This is reflected in our measure because the numerator is unlikely to significantly exceed the denominator in this case.
    \item If a source $s_i$ does not publish at time $t_i$, then $\resin_w^{(t_i, s_i)} = 0$. This means that $\embin_{w}^{(t_i, s_i)}$ will be identical to the global average at time $t_i$, and therefore $s_i$ can be neither a strong leader or follower.
    \item Suppose $w$ changes in meaning between time $t_1$ and $t_2$, but both $s_1$ and $s_2$ reach the new meaning in synchrony, as may happen if the change in meaning is caused by a sudden external event. In this case, $s_1$ and $s_2$ would both use an older sense of the word at $t_1$ and then switch to the newer sense at $t_2$. The numerator and denominator in~\autoref{eq:lead-equation} will be approximately equal, resulting in a leadership measure close to $1$.

\end{itemize}

We focus on the changes that took place over adjacent time periods, which are most likely to be indicative of inter-newspaper transmission. For each word in each successive temporal interval over the entire timespan, we calculated the lead of each newspaper over each of the others. We identified the two newspapers (one leader, one follower) that produced the maximum lead score and retained the pair if it passed the threshold obtained from the randomization procedure described below.

\subsection{Controlling for random noise and dataset artifacts}
\label{sec:method-control-randomness}
In any finite dataset, there will be spurious correlations, which may appear to be meaningful due only to random noise. In our dataset, there is added risk due to two sources of heterogeneity of the data. As shown in \autoref{fig:temporal-dist}, some newspapers publish towards the beginning of the dataset (from 1831) and others towards the end (1861-1865); furthermore, the corpus overall contains more text published in the years leading up to the Civil War. Second, the number and length of the articles published by each newspaper are widely divergent. While this heterogeneity reflects the reality of the historical trajectory of newspaper publishing at the time \citep[e.g.][]{Leonard_1995,Gross_Kelley_2010}, it is potentially problematic:
a statistical analysis of temporal trends might inherently focus on the newspapers that publish early and often as leaders, simply due to temporal precedence; similarly, newspapers that publish in greater volume may play an outsize role in determining the temporal word embeddings $\embin_{w_j}^{(t_i)}$, so that our assessment of semantic leadership will be most sensitive to changes in the words that these newspapers emphasize. 


While these issues can be partially mitigated by limiting the number of tokens per newspaper at each time step, additional controls are necessary due to the temporal heterogeneity of the data. We use randomization to control for both structural heterogeneity and random noise. 
We create a set of $K=100$ randomized datasets, in which word tokens are randomly swapped between newspapers. This preserves the structure of the dataset --- each newspaper has the same number of tokens within each time period --- but it breaks the link between individual newspapers and contextual word statistics. Any leadership relationships that are detected in such a randomized dataset must be attributed to either structural bias or random noise, because the word tokens have been assigned to individual newspapers at random. 

To see how these randomized datasets help us to control for noise and heterogeneity, recall that a leadership event is a tuple $(\edge{s_1}{s_2}, w, t_1, t_2),$ where $s_1$ and $s_2$ are newspaper sources, $w$ is a word, and $t_1$ and $t_2$ are timestamps. For each such event, we compute the leadership statistic from \autoref{eq:lead-equation} in each randomized dataset, yielding a set of values 
$\{\lead^{(k)}(\edge{s_1}{s_2},w,t_1,t_2)\}$ 
for $k \in 1\ldots K$. We can then compare these values with the score that was observed in the original, nonrandomized dataset. Our final set of influence events is the set of tuples such that,
\begin{equation}
\lead(\edge{s_1}{s_2}, w, t_1, t_2) > \Phi_{.95}\left(\{\lead^{(k)}(\edge{s_1}{s_2},w,t_1,t_2)\}_{k=1}^K \right),
\end{equation}
where the function $\Phi_{.95}(S)$ selects the 95th percentile value of the set $S$; as a special case, $\Phi_{.5}(S)$ is the median of $S$.
This procedure is based on principles from statistical significance testing~\citep{degroot2011probability}, and allows us to approximately bound the probability of each leadership score arising by chance.\footnote{Similar procedures which involve permutation have been used in network regression. For example, see \cite{simpson2001qap}}

\subsection{Aggregated semantic influence network}
\label{sec:method-aggregation-and-centrality}
\newcommand{\Count}[0]{\ensuremath c}
To generalize from individual leadership events, we aggregate the events into a semantic leadership network among the newspapers. Specifically, we construct an edge-weighted network, $G=(S,E)$, where, as before, $S$ is the set of newspapers that form the nodes in the network, and every $e \in E$ is a weighted edge, denoted $e = (s_1 \rightarrow s_2, \Count_{12})$, with $\Count_{12}$ indicating the number of words for which $s_2$ leads $s_1$.

We interrogate the structure of this aggregate influence network using two graph-theoretic measures. 
The first is \emph{pagerank}, which computes the overall centrality of each node in the network~\citep{page1999pagerank}; however, as we will see, this single measure fails to distinguish three cases of interest: newspapers that usually lead, newspapers that usually follow, and newspapers that generally do not engage with the rest of the network. For this reason, we perform a more fine-grained analysis using the \emph{HITS} metrics of \emph{hubs} and \emph{authorities}~\citep{kleinberg1999authoritative}. While details of these algorithms can be found in network analysis textbooks~\citep[e.g.,][]{newman2018networks}, we give brief descriptions here.

\paragraph{Pagerank.} The principle behind pagerank is that the importance of a node is a function of the importance of the nodes that link to it (its ``backlinks''). 
In our context, each edge $\edge{s_1}{s_2}$ is weighted by the number of innovations in which $s_2$ led $s_1$, and then this weight is normalized by the sum of weights on edges from $s_1$. This means that a newspaper has high pagerank if it leads other high pagerank newspapers, and particularly if it is the sole (or main) leader of those newspapers. Mathematically, the pagerank of node $i$ is given by:
\newcommand{\pagerank}[0]{\ensuremath \textsc{Pagerank}}
\begin{equation}
    \label{eq:pagerank}
    \pagerank_i = \alpha \sum\limits_{j} 
    \frac{A_{ij}}{\sum_{k}A_{kj}} \pagerank_j + \beta,
\end{equation}
where $A_{ij}$ is the weight on the edge from $j$ to $i$ (number of words for which $i$ leads $j$), $\pagerank_j$ is the pagerank of node $j$, 
and $\alpha$ and $\beta$ are additional free parameters.\footnote{We set $\alpha=0.85$ and $\beta=0.15/|S|$, as is typical of many applications of pagerank~\citep[see][for a discussion]{berkhin2survey}. 
} This formulation is recursive: the pagerank of each node is computed in terms of the pageranks of the other nodes. 
The overall pagerank problem can be formulated as a system of linear equations (one for each node), and these equations can be solved simultaneously by factorization of the matrix $A$. 
Pagerank has been used in prior work in the digital humanities~\citep{jockers2012computing}, and is closely related to eigenvector centrality, which is used more widely in the social sciences~\citep{jackson2010social}.

\paragraph{HITS.} A limitation of pagerank is that it assigns only a single importance score to any newspaper, which fails to differentiate followers from newspapers that simply do not participate in ongoing semantic changes. We therefore turn to a more fine-grained analysis technique. HITS, which stands for hyperlink-induced topic search, is another linear algebra-based network centrality algorithm, which decomposes central nodes into two complementary groups, \term{authorities} and \term{hubs}. Authorities are those nodes that are pointed to by high-scoring hubs, and conversely, hubs are nodes that point to high-scoring authorities. In our context, high scores as an authority and hub indicate the importance of a newspaper as a leader and follower respectively.

Mathematically, the authority and the hub score for a node $i$ can be given by the paired equations,
\newcommand{\hub}[0]{\ensuremath \textsc{Hub}}
\newcommand{\authority}[0]{\ensuremath \textsc{Authority}}

\begin{align}
    \label{eq:hits}
    \authority_i &= \alpha \sum\limits_{j} A_{ij}\; \hub_{j}, \\
    \hub_i &= \beta \sum\limits_{j} A_{ji}\; \authority_{j},
\end{align}
where $A_{ij}$ is the weight on the edge from $j$ to $i$ (number of words for which $i$ leads $j$), $\authority_i$ is the authority score of node $i$, $\hub_i$ is the hub score of node $i$.
\footnote{Unlike pagerank, $\alpha$ and $\beta$ are not needed to be preset and are related to the leading eigen value of $A{^T}A$ and $AA{^T}$~\citep{newman2018networks}.} Similar to pagerank, HITS also originated with the application of ranking webpages in a hyperlinked environment, but it has been increasingly used in information sciences~\citep[e.g.][]{jiang2012towards} and digital humanities~\citep[e.g.][]{sudhahar2015network}.

\section{Data}
\label{sec:data}
We analyze a subset of nineteenth century newspapers digitized and hand-keyed by Accessible Archives.\footnote{https://www.accessible-archives.com/} We obtained permission to scrape the contents of several of Accessible Archives's collections, including "African American Newspapers," the "Women's Suffrage Collection," and several individual titles. Each newspaper in the archive consists of multiple issues, and each issue contains multiple articles. Each article is encoded as a unique \texttt{html} page. We extracted the text from each page using \texttt{beautifulsoup4} package in python, following a process similar to \cite{Klein_Eisenstein_Sun_2015}. We further process the data using the methods described below. 

\subsection{Contents and Metadata }
\label{sec:data-metadata}

\begin{figure}
    \centering
    \includegraphics[width=0.7\textwidth]{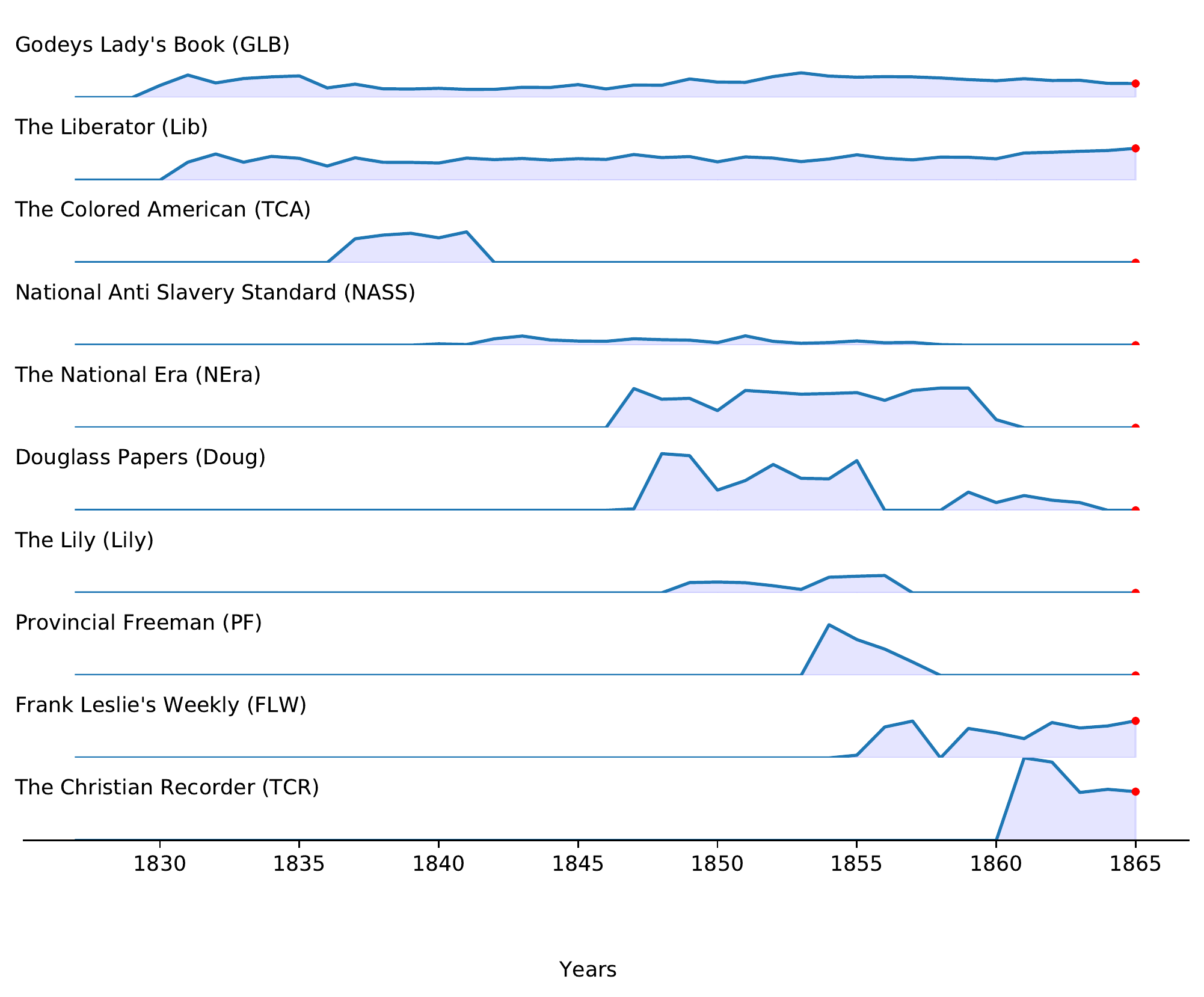}
    \caption{The distribution of data in Accessible archives, after further processing to remove duplication and fixing digital errors, that is used for analysis in the rest of the paper.}
    \label{fig:temporal-dist}
\end{figure}

For the purposes of this study, we limit our corpus to the period between March 1827, the date of the earliest newspaper in our collection, and December 1865, which marks the ratification of the Emancipation Proclamation. 
Most newspapers in our collection published weekly, though some titles are monthly publications. The temporal distribution of the newspapers in our collection is given in~\autoref{fig:temporal-dist}, and a detailed description of the newspapers is found in \autoref{tab:newspapers-description}. The collection is temporally skewed: the later years have more newspapers and more articles. This reflects a general trend in the rise in abolitionist newspapers published in the United States~\citep{Leonard_1995}, as well as the influence of scholarship on the subject, which has focused on specific titles published in the final decades of our study, leading to their being digitized more quickly and more completely than other less-known works~\citep{Fagan_2016}. 

\begin{table}[H]
    \footnotesize
    \centering
    \begin{tabular}{p{0.2\textwidth}lp{0.65\textwidth}}
        \toprule
        Title & First Issue & Description \\
        \midrule
        Freedom’s Journal & 1827 & An abolitionist newspaper established by two Black clergymen, Samuel E. Cornish and John Brown Russworm. The earliest newspaper edited by and for Black Americans presently known. \\
        The Liberator & 1831 & An abolitionist newspaper established by William Lloyd Garrison, the famed white abolitionist. It was known for its fiery rhetoric and is considered among the most influential papers of its day. \\
        The Colored American & 1837 & An abolitionist newspaper established by Phillip A. Bell as \textit{The Weekly Advocate}. Samuel Cornish, of \textit{Freedom's Journal}, became its editor several months later, and initiated the name change. \textit{The Colored American} was intended to serve as a national publication edited by and for Black Americans. \\
        National Anti-Slavery Standard & 1840 & The official newspaper of the majority-white American Anti-Slavery Society. The author and abolitionist Lydia Maria Child was its first editor. It was intended to serve as a more moderate counterpart to \textit{The Liberator}, and appeal to a white reading public. \\
        The Douglass Papers & 1847 & A set of abolitionist newspapers edited Frederick Douglass, the author and orator who liberated himself from slavery. \textit{The North Star} was launched in 1847, with Martin Delaney as co-editor. Douglass renamed it \textit{Frederick Douglass's Paper} upon Delaney's departure in 1851. After a several year pause, Douglass launched \textit{Douglass's Monthly} in 1859. \\
        The National Era & 1847 & A more general newspaper, founded by Dr. Gamaliel Bailey, Jr. and aimed at a white audience. It supported the cause of abolition, and is most famous for being the first to publish \textit{Uncle Tom's Cabin}. \\
        The Christian Recorder & 1854 & A newspaper established by the African Methodist Episcopal (AME) Church. Once overlooked for its contributions to abolition, it is now understood as central to the movement for Black liberation in the nineteenth century. \\
        Provincial Freeman & 1854 & An abolitionist newspaper established by Mary Ann Shadd, a Black abolitionist, educator, and later lawyer, who emigrated from the US to Canada. The \textit{Freeman} is believed to be the first newspaper edited by a Black woman in all of North America. \\
        The Lily & 1849 & The first known women's suffrage newspaper founded by Amelia Bloomer, who was white. Its initial emphasis was on the temperance movement, but it became more invested in women's rights over time. \\
        Godey’s Lady’s Book & 1830 & A monthly women's magazine edited by a white woman, Sarah Josepha Hale, for the majority of its run. It was rarely overtly political. It is included in the corpus as a point of comparison with \textit{The Lily}. \\
        Frank Leslie’s Weekly & 1855 & A weekly magazine aimed at a white reading public. It offered a wide range of content, and is best known for its literary selections and coverage of current events. It is included in this corpus as a point of comparison to both the abolitionist and suffragist papers of the same era. \\
        \bottomrule\\
    \end{tabular}
    \caption{\textbf{Detailed description of all newspaper titles}. We include \textit{Freedom's Journal }in identifying semantic changes but exclude it from the subsequent analysis because its publication stopped very early in the timeframe that we considered.}
    \label{tab:newspapers-description}
\end{table}

The focus of this study is on the abolitionist movement of the nineteenth-century United States. This is reflected in the titles included in our corpus, the majority of which explicitly identify as abolitionist newspapers. Our corpus is notable for its relatively equal number of Black-edited and white-edited abolitionist titles; this was among the reasons we selected Accessible Archives as our original data source. Accessible Archives also provides information about the race and gender of the editors of each newspaper, as well as about the paper's intended audience; we confirmed the identity of each editor and each newspaper's audience with a second scholarly source, and then converted this information into additional metadata for use in the project. During this period, most newspapers were associated with a small group of editors (on the order of one to three people), which remained relatively constant over the newspaper's run, and who exerted a high degree of control over the newspaper's contents~\citep{Casey_2017}. 

For this project, we supplemented the abolitionist titles with additional newspapers linked to the women's suffrage movement, so as to enable a comparison between the two movements, which were often linked. We have also included a more general newspaper of that same era, as well as two monthly magazines. Our rationale for including these titles was to provide a baseline for any claims about abolition (or women's rights) that we might make. While this dataset is far from exhaustive, it represents a meaningful ``scholarly edition of a literary system,'' as Katherine Bode might describe it, one which captures some of the key titles in the scholarship on the subject along with known interlocutors and additional likely influences \citep{Bode_2018}. For more on the history of these newspapers, see \cite{Gross_Kelley_2010}. 


\subsection{Data Processing}
\paragraph{Digital error correction.} While Accessible Archives' hand-keyed texts are far less error-prone than those digitized using optical character recognition(OCR)~\citep{smith2007overview}, we discovered several types of errors in the course of our analysis. One common source of error in this collection is whitespace.\footnote{Other digital errors (e.g. first or last letter of a word getting stripped) also occur in the corpus but are not systematic.} Such errors, also known as \emph{word segmentation errors} or \emph{spacing errors}, arise during both OCR and the post-digitization handling of the data~\citep{kissos2016ocr} --- the latter being the source of the error in the Accessible Archives corpus --- and result in the elimination of whitespace between words. This leads to out-of-vocabulary items like \example{senatoradmits} and \example{endowedwith}. Our approach to correcting these errors is described in prior work \citep{soni2019correcting}. 

\paragraph{Deduplication.} The collection also contains a number of articles that were reprinted verbatim from other newspapers (e.g. mission statements, notices, speeches) as well as many that were reprinted with minor modifications (e.g. advertisement campaigns, commodity price reports). While these reprinted articles are an artifact of the time~\citep[see][]{Cordell_Smith_Mullen_Fitzgerald_Forthcoming}, their presence in the corpus is problematic for modeling, potentially resulting in false positive semantic changes. As a conservative heuristic, we remove from the corpus all articles that share a contiguous sequence of eight or more words with another article. Though this heuristic can also remove a number of non-duplicate articles, we apply it because it does not affect the distribution of the documents temporally or across newspapers, and keeps roughly 90,000 articles in the corpus. 

\paragraph{Data and experimental setup.} For all the analysis, we divide the corpus into ten equal time intervals.\footnote{With $5$ and $20$ time intervals, the discovered semantic changes were very similar, with approximately 90\% of all changes preserved across all three settings. However, increasing the number of intervals made the residuals more difficult to compute, while decreasing the number of intervals made it difficult to isolate the precise time of changes. Consequently, we set the number of intervals to ten as a tradeoff.} Newspapers that do not contain at least $500$ articles throughout their tenure in the period of interest are ignored. To maximize the number of newspapers that meet this condition, we group together several smaller newspapers that share an editor. Namely, we consolidate \newspaper{The North Star}, \newspaper{Frederick Douglass's Paper}, and the \newspaper{Douglass Monthly} into a single source we label \newspaper{Douglass Papers}, because each was edited by the Black abolitionist and orator Frederick Douglass; we also group \newspaper{The Weekly Advocate} with \newspaper{The Colored American}, since the former changed its name to the latter several months into its publication run. 

\paragraph{Vocabulary and hyperparameters.}
All the text is lowercased before processing. The vocabulary is limited to the $50,000$ most frequent words in the corpus. The embedding size for all the models is set to $100$ dimensions. The window size to determine the context words is set to $5$ in both directions. The $\ell_2$ regularization penalty is set to $1e^{-4}$.
\section{Results}
\label{sec:results}

Our method offers insights about the evolution of conceptual terms related to abolition, as well as about the specific newspapers that contributed to the spread of the terms' changed meanings. By considering these terms in relation to others identified by the model as associated with particular newspapers, we elaborate an argument about the nature of semantic leadership and its relation to social and political influence within our newspaper corpus. We also discuss the significance of the overall influence network generated by our model, which confirms aspects of our previous understanding of the movement's intellectual leaders and followers, while illuminating additional pathways of influence among some of the understudied newspapers in our corpus. Taken together, these results help to establish the validity of our method and, we hope, will seed new research questions about these important newspapers, the relationships among them, and their significance for our understanding of the abolitionist movement.  We now discuss the findings that result from each phase of our methodological pipeline, building towards the conclusion about women's leadership and the role of the Black press. 

\subsection{Semantic changes}
\label{sec:results-semantic-changes}

\begin{table}
    \footnotesize
    \centering
    \begin{tabular}{lp{0.1\textwidth}p{0.1\textwidth}p{0.3\textwidth}p{0.3\textwidth}}
    \toprule
    Word & From & Until & Earlier examples & Later examples\\
    \midrule
    \example{cabinet} & $1831$-$1835$ & $1861$-$1865$ & \example{a \textbf{cabinet} of antiquities} & \example{president and his \textbf{cabinet}}\\
    & & & \example{the secret \textbf{cabinet} on the right} & \example{members of the \textbf{cabinet}}\\[1ex]
    \example{equality} & $1827$-$1831$ & $1853$-$1857$ & \example{liberty and \textbf{equality} to all} & \example{the \textbf{equality} of the races}\\
    & & & \example{blessed with liberty and \textbf{equality}} & \example{an \textbf{equality} with the whites} \\[1ex]
    \example{freedom} & $1827$-$1831$ & $1857$-$1861$ & \example{the spirit of \textbf{freedom} is marching}& \example{opposition of slavery to \textbf{freedom} continues}\\
    & & & \example{establish their \textbf{freedom} from being sold} & \example{colored man for \textbf{freedom} and self-government}\\[1ex]
    \example{rights} & $1841$-$1845$ & $1861$-$1865$ & \example{inalienable \textbf{rights} of man} & \example{advocate the \textbf{rights} of the black man} \\
    & & & & \example{the \textbf{rights} of women} \\[1ex]
    \example{justice} & $1831$-$1835$ & $1861$-$1865$ & \example{legislature or the court of \textbf{justice}} & \example{one simple act of \textbf{justice} to the slave}\\
    & & & \example{james madison and chief \textbf{justice} marshall} & \example{emancipation as an act of \textbf{justice} and humanity}\\[1ex]
    \example{immediate} & $1827$-$1831$ & $1849$-$1853$ & \example{the \textbf{immediate} legislation of congress} & \example{hostile to \textbf{immediate} emancipation} \\
    & & & \example{the \textbf{immediate} extinction of slavery} & 
    \example{safety of \textbf{immediate} unconditional emancipation} \\[1ex]
    \example{fight} & $1827$-$1831$ & $1849$-$1853$ & \example{you are paid to \textbf{fight}} & \example{profess to \textbf{fight} for liberty} \\
    & & & \example{a \textbf{fight} took place} & \example{determination to \textbf{fight} till the last} \\[1ex]
    \example{service} & $1831$-$1835$ & $1861$-$1865$ & \example{first military \textbf{service} was in corsica} & \example{gallant \textbf{service} at fort wagner} \\
    & & & \example{their \textbf{service} in the army} & \example{\textbf{service} was held in the church} \\[1ex]
    \example{aid} & $1831$-$1835$ & $1861$-$1865$ & \example{offers of \textbf{aid} from colonizationists} & \example{state fugitive \textbf{aid} society} \\
    & & & \example{gave them \textbf{aid} and tyrants fled} & \example{liberal offers to \textbf{aid} the cause of freedom}\\[1ex]
    \example{growing} & $1835$-$1839$ & $1861$-$1865$ & \example{\textbf{growing} up in shame and poverty}& \example{republic is \textbf{growing} darker everyday}\\
    & & & \example{extreme want \textbf{growing} out of the toil} & \example{stole the \textbf{growing} light of dawn}\\[1ex]
    \example{writing} & $1827$-$1831$ & $1853$-$1857$ & \example{coloured adults in reading, \textbf{writing}, arithmetic} & \example{the \textbf{writing} is blotted in many places}\\
    & & & \example{reading or \textbf{writing} the letters seem} & \example{full of \textbf{writing} in a round-text hand}\\[1ex]
    \example{hoped} & $1841$-$1845$ & $1845$-$1849$ & \example{it is \textbf{hoped} that every newspaper} & \example{promised and \textbf{hoped} for but homeless future} \\
    & & & \example{it was \textbf{hoped} however that the friends of temperance} & \example{he now \textbf{hoped} to have peace}\\[1ex]
    \example{courage} & $1827$-$1831$ & $1849$-$1853$ & \example{possessed of both sense and \textbf{courage}} & \example{renewed \textbf{courage} instead of the hopeless feeling of banishment} \\
    & & & \example{wit and \textbf{courage} amongst all} & \example{personal \textbf{courage} and devotion}\\
    \bottomrule\\
    \end{tabular}
    \caption{\textbf{Examples of semantic changes}. The terms are detected to have changed most in meaning from the earlier timespan to the later.}
    \label{tab:semantic-changes-examples}
\end{table}

The first step in our method identifies terms that undergo semantic change. Illustrative examples are documented in \autoref{tab:semantic-changes-examples}. The word \example{equality}, in its earliest usage in the corpus (1827-1831), seems to express the idea of equality in a Lockean sense. It is nearest in meaning to other key terms in the Enlightenment discourse of natural rights, such as \example{liberty}, \example{rights}, \example{wisdom}, and \example{mankind} --- the same terms that were commonplace at the time of the nation's founding, and into the early years of the republic.  Over time, however, the usage of the term becomes more connected to questions of the practice of democracy. Although it retains some of its original associations, by $1857$ \example{equality} also becomes closely associated with the idea of \example{self-government} issues of \example{suffrage}, and the specific \example{guarantees} of the ideals expressed in nation's founding documents. While this shift from abstract to concrete is often attributed to the authors of the nation's founding documents, as they attempted to translate political philosophy into a governing structure, it is interesting to observe that in the abolitionist press (and related publications), the shift happens several decades later. Could it reflect a concerted effort, on the part of the abolitionist press, to first frame the issue of slavery in philosophical terms before recognizing that slavery's abolition might require a more concrete argument? Could it reflect the growing influence of the women's suffrage movement, which seized upon equality as its governing ideological principle? While the changing meaning of \example{equality} does not offer a definitive answer to these questions, it helps point to how the changing meanings of individual words, as documented in our corpus, can index larger ideological changes and debates.

The term \example{freedom} follows a similar, if more chronologically consistent shift from abstract to concrete. It enters the corpus with general associations to the nation's foundational ideals. Its near neighbors include words like \example{humanity}, \example{people}, and \example{country}. By the time of beginning of the Civil War, however, the associations of the term have become more specific, including references to both \example{liberty} and \example{slavery}, as well as terms that reference arguments about the \example{rights} and \example{guarantees} of the nation that should ensure freedom, as well as the \example{institutions} that should enforce it. With respect to the term \example{rights}, we find that this term also moves from a more abstract to a more concrete meaning, tracking both \example{freedom} and \example{equality} --- terms which appear in its list of near neighbors, along with terms like  \example{humanity}, which reinforce a sense of the term that is broadly applied. Over the second half of the corpus, however, the term narrows in its application to questions of \example{citizenship} and \example{suffrage}. 

As with the questions prompted by the term \example{equality}, knowledge of which newspaper was most responsible for this narrowed sense of the term might tell us something about who was responsible for shifting the conversation about \example{rights} and their application. Was it one of the newspapers edited by women? This finding would help to confirm the argument made by Manisha Sinha, among others, that women abolitionists were responsible for pushing the movement to consider issues of enfranchisement and women's rights more broadly. Or was it one of the papers associated with the Black press? This might introduce another facet of the argument made by Derrick Spires about how Black Americans, both before and after emancipation, understood citizenship as a broad concept, one that transcended any legal definition in order to incorporate civic engagement, public expression, and mutual aid. 

A final term, \example{justice}, is worth exploring in this context. Like \example{freedom} and \example{rights}, this term also follows a trajectory that is consistent with historical scholarship on the subject, albeit one that moves in the reverse direction from concrete to abstract. It enters the corpus with narrow associations to the legal system, as evidenced by near neighbors including \example{judges}, \example{trial}, \example{offense}, and \example{crime}. But by the final years of the corpus, the term has significantly expanded, commanding a more ideological frame. Its associations at the onset of the Civil War include the terms \example{liberty}, \example{equality}, \example{rights}, and \example{oppression}, which tracks arguments about how justice transformed into a much more powerful concept over the course of the nineteenth century as a result of the criminal justice movement. In our corpus, this transformation is reinforced by the appearance of the terms \example{universal}, \example{humanity}, \example{citizenship}, and \example{nation}, which suggest the success of those who advocated for criminal justice in expanding not only legal protections, but also ideas about what \example{justice} properly entailed. 

Taken together, these terms demonstrate how changes in near neighbors, coupled with information about which newspapers were most responsible for those changes, can point to a new understanding of (or, alternately, enhance existing knowledge about) the abolitionist movement and its conceptual undercurrents. 

\subsection{Semantic Leadership}
\label{sec:results-semantic-leaders}

Of the changes discussed in the section above, two --- \example{justice} and \example{rights} --- were attributable to specific newspapers. In terms of \example{justice}, interestingly, the newspaper most closely associated with the more ideological conception of the word was not an abolitionist newspaper at all, but rather, the women's suffrage newspaper, \newspaper{The Lily}. More interesting still, it is \newspaper{The Liberator} --- the long-running abolitionist newspaper with a reputation for its impassioned arguments against slavery --- which is identified as the most significant follower in this new usage. Could it be that \newspaper{The Liberator}, edited by a white abolitionist, William Lloyd Garrison, was not as ideologically innovative as the historical narrative would lead us to believe? We continue this line of inquiry below.  

As for \example{rights}, our method identifies the two more general newspapers, both intended for a white readership, as the leader-follower pair. In the period between 1853 and 1861, \newspaper{The National Era} is the most responsible for narrowing the scope of the conversation about rights from abstract and ideological to legal and concrete; and it is \newspaper{Frank Leslie's Weekly}, a weekly magazine which also published a wide range of content, which most closely followed this trend. That the conversation about rights is bounded by white, general-audience newspapers suggests that this concept, with its anchor in political philosophy, may hold less relevance to the abolitionist press, or to women's suffrage newspapers, than concepts involving ideas about morality, humanity, or other more expansive understandings about the implications of the persistence of slavery. 


Beyond the analysis of individual words and the newspapers that lead in their changing meanings, it is also possible to examine any two newspapers as leader-follower pairs. Comparing two newspapers that published at the same time as each other, and were known to be in conversation --- for example, \newspaper{The Colored American}, a newspaper edited by a Black editorial staff which circulated among a predominantly Black audience, and \newspaper{The Liberator}, mentioned above, edited by William Lloyd Garrison, who was white, and which circulated among a mixed, if predominantly white audience --- we find that \newspaper{The Colored American} leads on the terms \example{immediate} and \example{fight}. While not explicitly political terms, they do suggest a tone of urgency that might surround an argument for liberation, one which helps to further dismantle the narrative of \newspaper{The Liberator} as the most radical of the abolitionist papers, and in the process, provides additional evidence to support a claim that \newspaper{The Colored American}, led by Black abolitionists, deserves more of the credit for accelerating the fight against slavery. 

When considering how newspaper pairs and the words that connect them can open up new research questions, we might consider the leader-follower relationship between \newspaper{The Christian Recorder}, the official newspaper of the African Methodist Episcopal (AME) Church, and \newspaper{Godey's Lady's Book}, a magazine aimed at white women. \newspaper{The Christian Recorder} is among the several publications of the AME Church, including \newspaper{The Christian Recorder} that have been posited as key sources in the emergence of a Black print culture~\citep{Foster_2005, Fagan_2016}, and yet \newspaper{The Christian Recorder} stands far to the side in our influence network, neither leading nor following many other newspapers. This suggests that the conversations taking place in that paper were disconnected from the others --- in terms of language and therefore abolitionist thought. A prior thematic analysis confirms that the articles did indeed consist mostly of religious content~\citep{Klein_2020}. Yet our model detects that the paper \emph{does} lead the rest of the network on the terms \example{service} and \example{aid}. These terms clearly relate to core tenets of Christianity, but is interesting to note that the main follower is not another abolitionist paper but instead a white women's magazine. This suggests an unexpected new line of inquiry: how Christian benevolence, often framed as the contribution of white women's involvement in the abolitionist movement, was instead rooted in the Black church.

\begin{figure}
    \centering
    \includegraphics[width=0.8\textwidth]{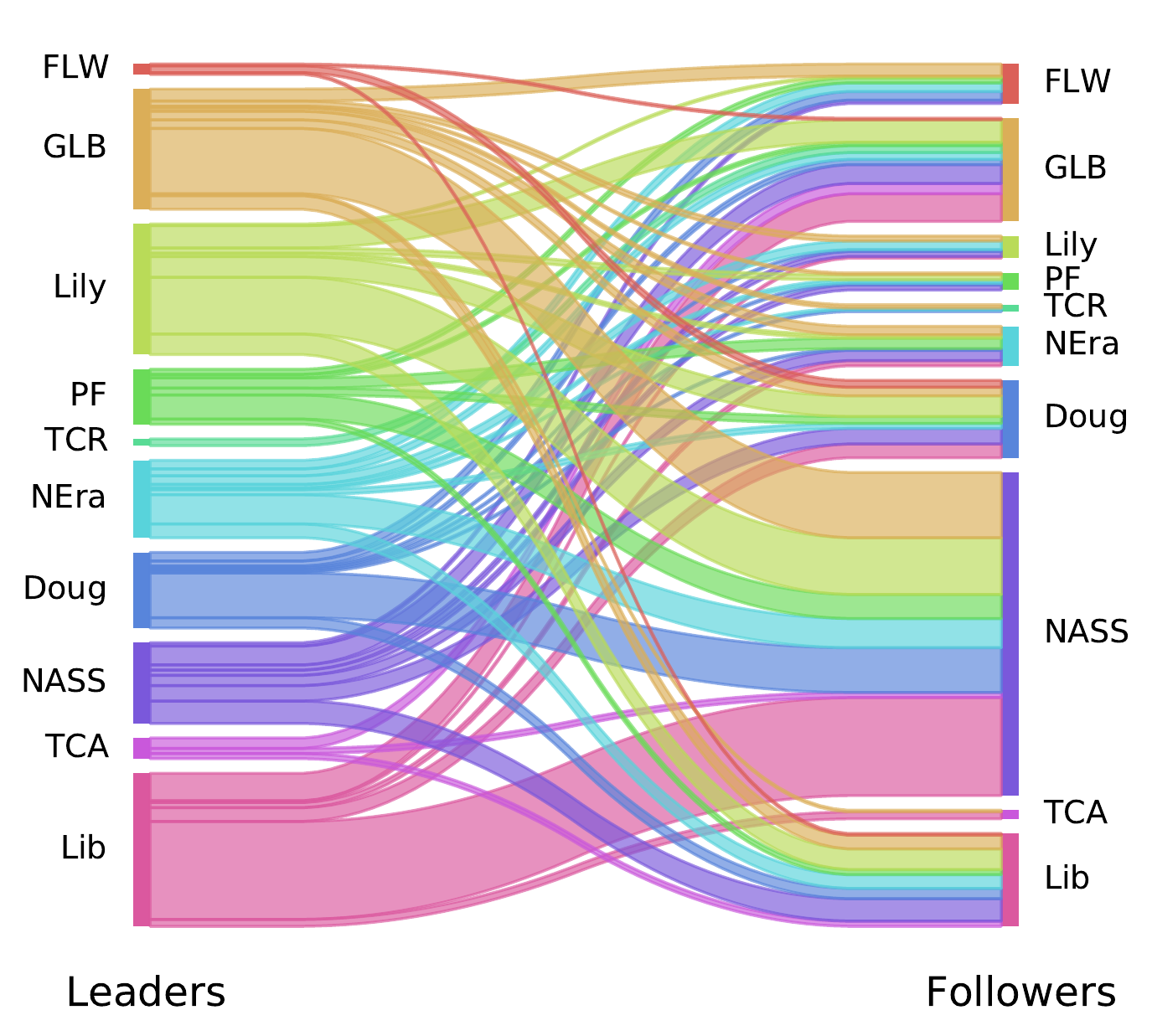}
    \caption{\textbf{The leader-follower newspaper pairs}. The leading newspapers are shown on the left and the trailing newspapers on the right. Each newspaper is shown as a rectangular patch with height proportional to the number of words for which it is considered a leader (or follower). The thickness of each stripe connecting two newspapers is proportional to the number of words between the newspapers that the stripe connects.}
    \label{fig:network-sankey-main}
\end{figure}

Thus far, we have considered individual newspaper pairs and the specific words which connect them. But we can also analyze these words in aggregate. Our method identifies 435 semantic leadership events, containing a mixture of political terms and activist language, as well as other words that are not easily connected to any particular political or ideological stance. In fact, words like \example{growing} and \example{writing}, \example{hoped} and \example{courage} --- words which suggest generalized movement, motion, action, and emotion --- emerged with some of the highest leadership scores. This finding suggests that the influence detected by our model might be better understood at the level of what might be described as \emph{discourse}, rather than at the level of individual words. 

\begin{figure}
  \begin{subfigure}[t]{.49\textwidth}
    \centering
    \includegraphics[width=\linewidth]{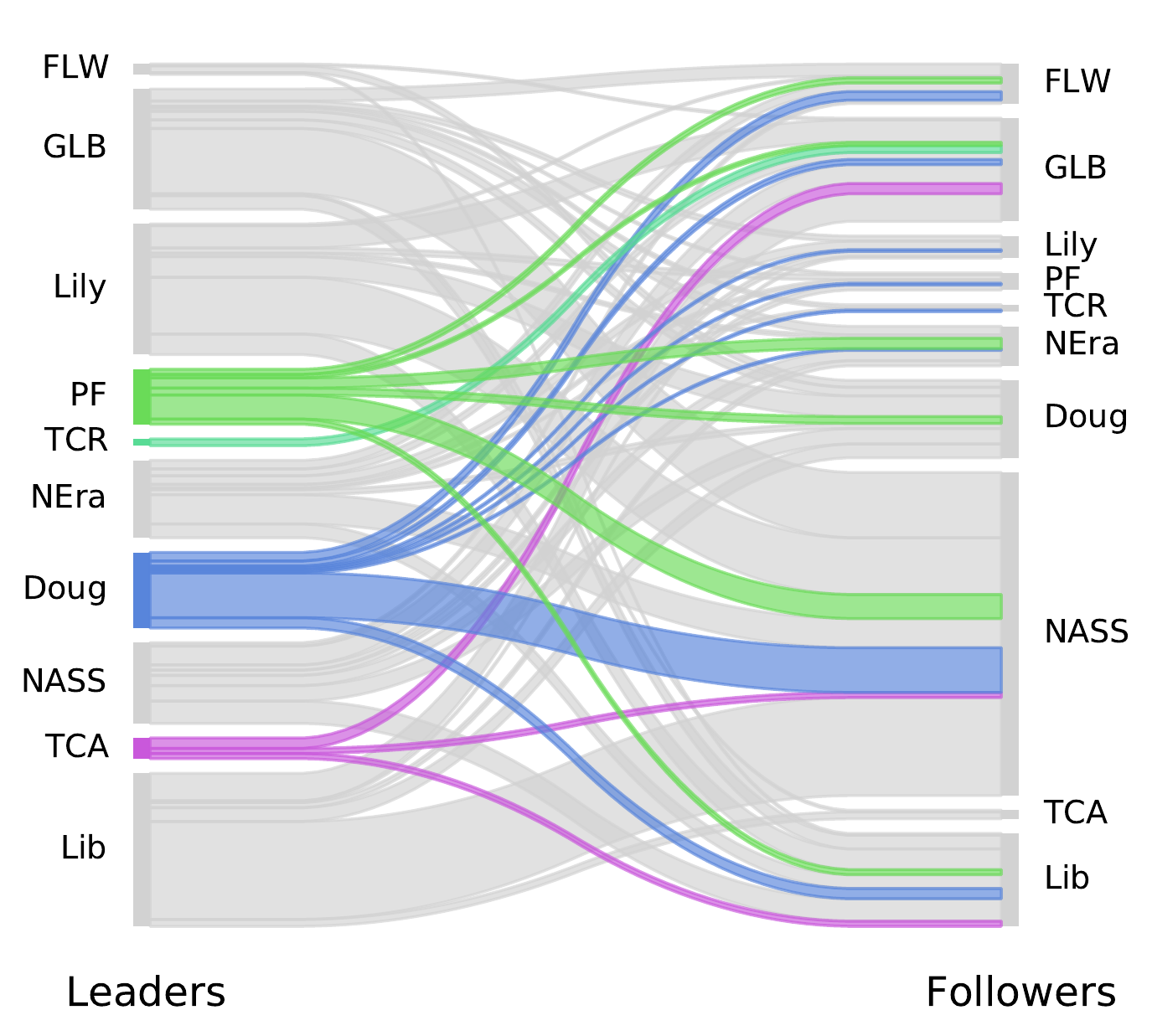}
    \caption{Newspapers with Black editors as leaders}
  \end{subfigure}
  \hfill
  \begin{subfigure}[t]{.49\textwidth}
    \centering
    \includegraphics[width=\linewidth]{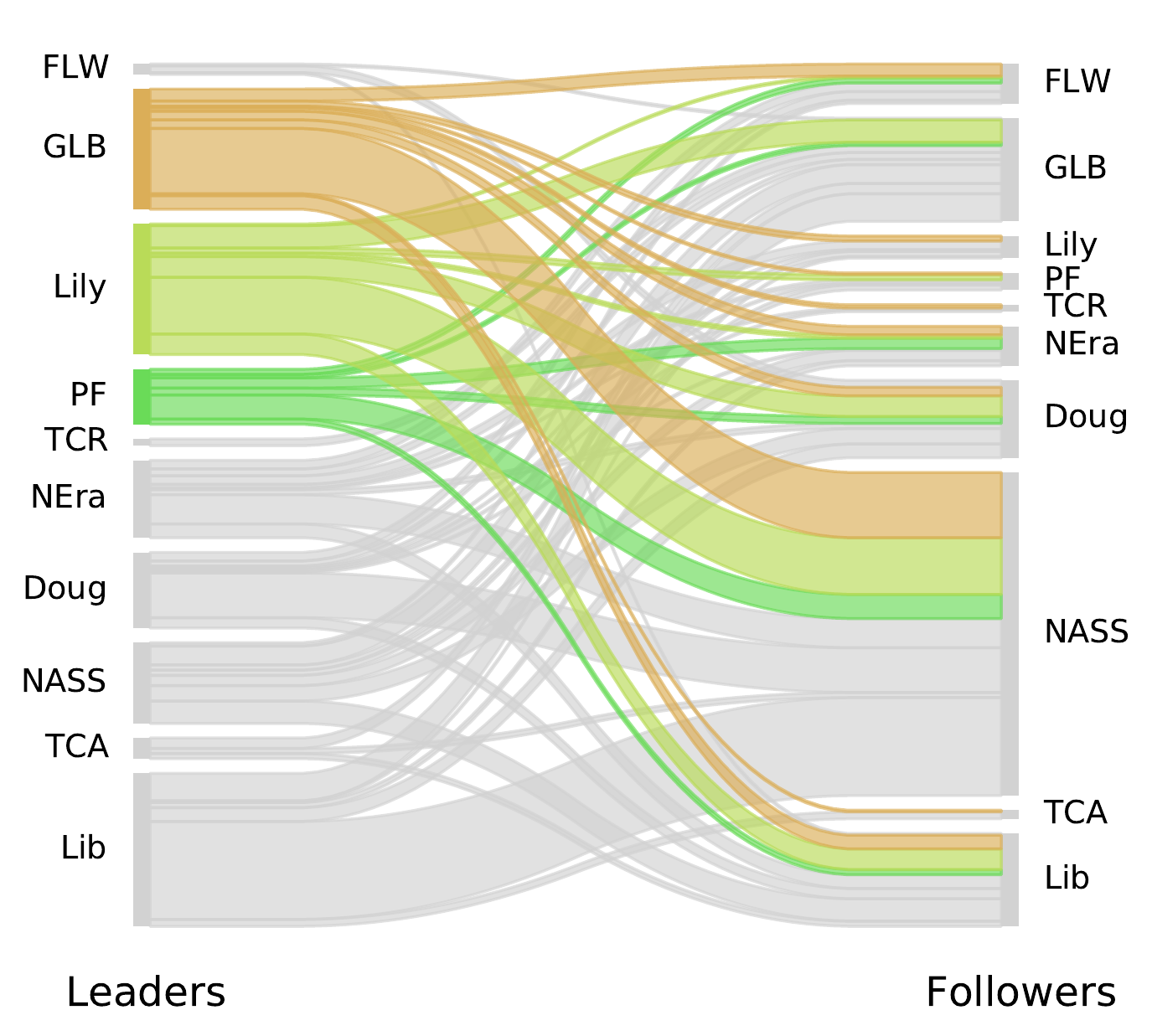}
    \caption{Newspapers with women editors as leaders}
  \end{subfigure}

  \medskip

  \begin{subfigure}[t]{.49\textwidth}
    \centering
    \includegraphics[width=\linewidth]{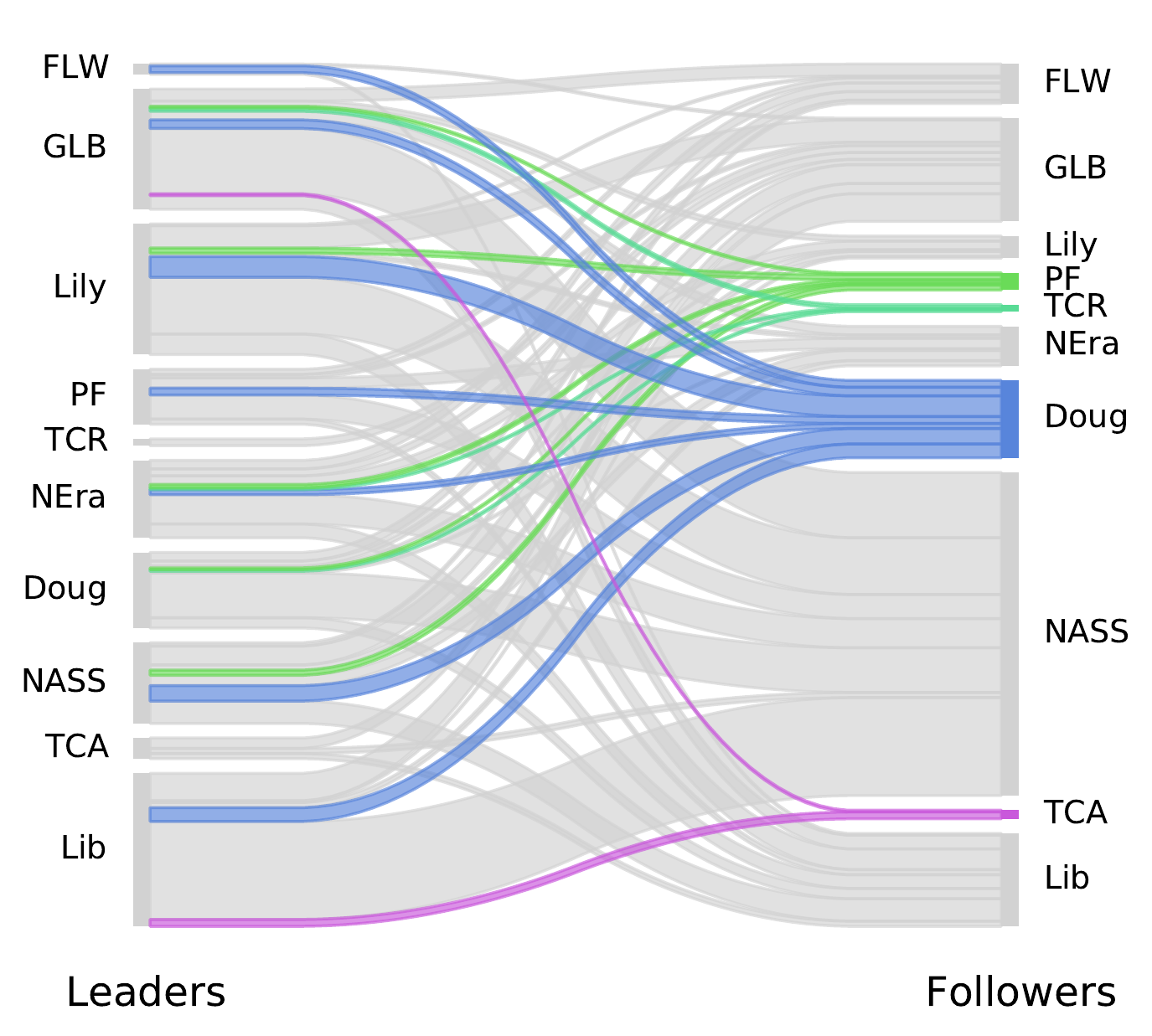}
    \caption{Newspapers with Black editors as followers}
  \end{subfigure}
  \hfill
  \begin{subfigure}[t]{.49\textwidth}
    \centering
    \includegraphics[width=\linewidth]{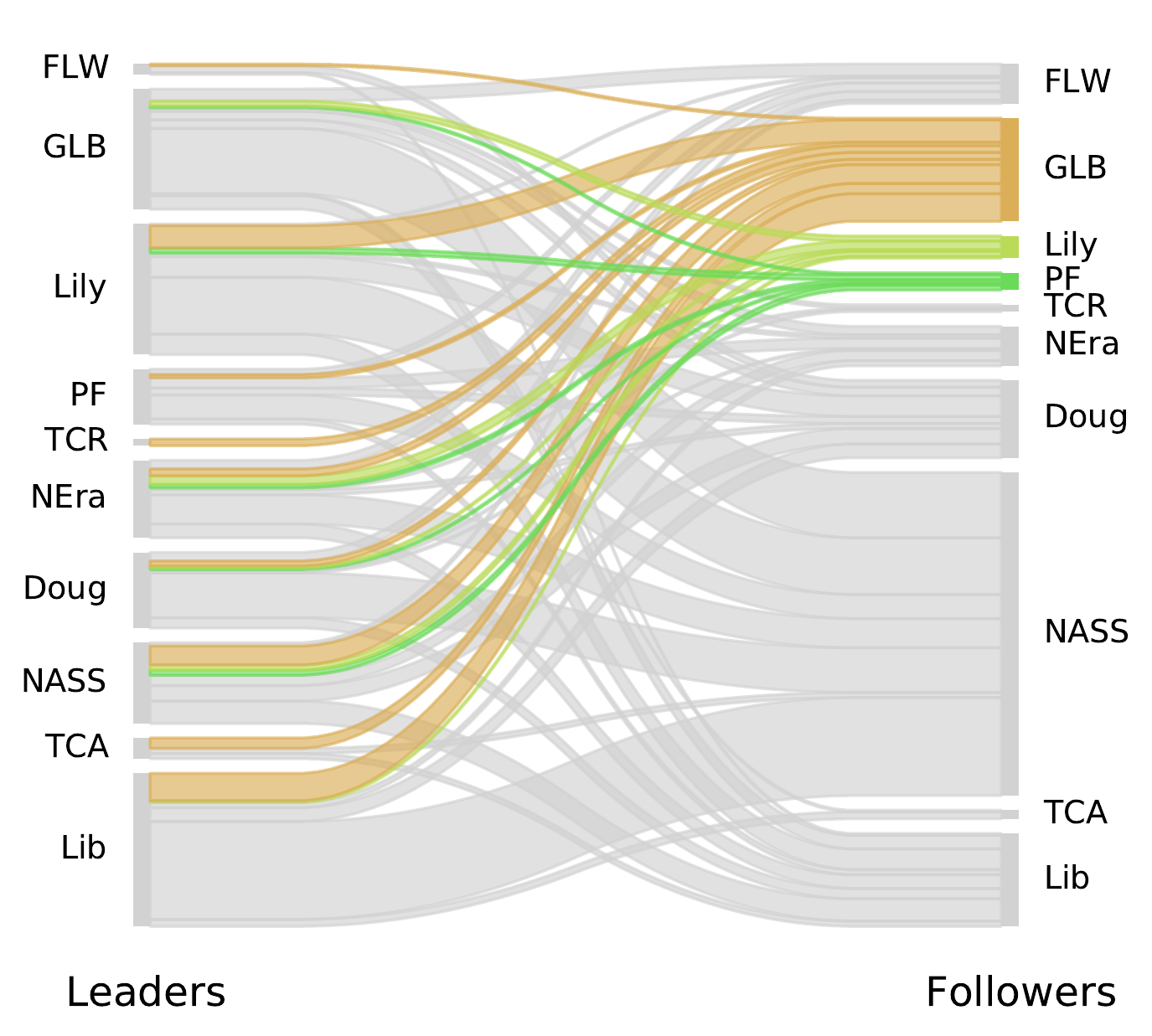}
    \caption{Newspapers with women editors as followers}
  \end{subfigure}
  \caption{\textbf{Focused lead-lag relationships} We highlight the dyads in which the newspapers that had Black editors (left) and women editors (right). The top row shows the dyads in which these newspapers lead, while the bottom row shows the dyads in which they trail. For more details on the editorial coding of newspapers, see \nameref{sec:data-metadata}.}
  \label{fig:network-sankey-editor-highlight}
\end{figure}

Looking at the aggregated counts of leader-follower pairs confirms this hypothesis: the relationship between \newspaper{The Liberator} and \newspaper{The National Anti-Slavery Standard} rises to the top of the list, with 32 words (of the 435) on which \newspaper{The Liberator} leads and \newspaper{The National Anti-Slavery Standard} follows. This is a valuable if unsurprising result, since of all of the newspapers in our corpus, historical circumstance suggests these two titles \textit{should} be in close dialogue with each other, as they were both published by the American Anti-Slavery Society. The relationship with the second highest count, between \newspaper{Godey's Lady's Book} and the \newspaper{Standard}, is more surprising: the \newspaper{Standard} was launched in no small part to bring women around to the abolitionist cause, and a woman --- the white abolitionist and author Lydia Maria Child --- edited the paper in its early years. The topical analysis performed by \cite{Klein_2020} revealed that the \newspaper{Standard} did indeed contain more thematic content related to women, but the identification of the \newspaper{Standard} as a follower of \newspaper{Godey's Lady's Book} provides new evidence about the centrality of women's issues to \newspaper{Standard}. Intriguingly, the \newspaper{Standard} also follows \newspaper{The Provincial Freeman}, an abolitionist newspaper edited by a Black woman, in outsized degree to the \newspaper{Freeman}'s far smaller circulation. Edited by Mary Ann Shadd (later Carey), the \newspaper{Freeman} was known, like \newspaper{The Liberator}, for its uncompromising editorial tone \citep{Rhodes_1998,Casey_2019}. As a result, it struggled to find a wide readership \citep{Rhodes_1998}. It was also published out of western Canada, which further limited its reach. And yet here we find its influence reaching quite far indeed, through the pages of the widely-circulated \newspaper{National Anti-Slavery Standard}, which adopted aspects of its discourse. Notably, the \newspaper{Freeman} also holds sway over the newspapers edited by Frederick Douglass, which have dominated accounts of the Black press. Taken together, these aggregated leader-follower pairs point to how a quantitative analysis conducted at the level of changes in individual words, and subsequently subjected to an analysis in terms of leadership, can help shift the narrative about the abolition and those responsible for shaping its discourse.

\subsection{Semantic Leadership Network}
\label{sec:results-network}

It is also possible to aggregate the leader-follower pairs and their changes all together in order to form a directed, weighted network. In \autoref{fig:network-sankey-main}, we display leader-follower the relationships among all of the newspapers considered in our study. In this diagram, all of the newspapers in our corpus are listed in order on both sides. A chord that emerges from a newspaper's name on the left indicates that it leads on a single word, while one that connects on the right indicates that the newspaper follows on that term. Thus, the vertical length next to any particular title indicates the number of words that the newspaper leads on or follows, respectively. Examining the titles associated with the highest number of leading and following words, we find confirmation of two known findings: first, that \newspaper{The Liberator} holds broad influence over many of the other newspapers, as intimated above; and second, that the \newspaper{National Anti-Slavery Standard} tends to follow much more than it leads. This reflects the rationale for launching the paper, which was to provide a more moderate publication than \newspaper{The Liberator} so as to appeal to a broader coalition of potential supporters. (Recall that both newspapers were published by the \newspaper{American Anti-Slavery Society}.)

Just below those top-level leaders and followers, the most prominent leaders are \newspaper{The Lily} and \newspaper{Godey's Lady's Book}. Both newspapers (mostly) edited by and written for white women; neither is an abolitionist newspaper, so it is surprising that they are found to lead in a corpus are centered on abolition. While more research is required in order to understand the source of this influence, we can learn a bit more by examining the relationships among them, as pictured in \autoref{fig:network-sankey-editor-highlight}. Here we see two things: first, that the influence of these publications can be detected in every single newspaper in our corpus; and second, that \newspaper{The National Anti-Slavery Standard} and Frederick Douglass's various papers claim the majority of their influences from these women's publications.

When considering the sub-network of newspapers associated with the Black press (\autoref{fig:network-sankey-editor-highlight}), we again observe that the influence of these papers is distributed across the corpus. Two newspapers dominate this sub-network: Frederick Douglass's newspapers and \newspaper{The Provincial Freeman}. But it is worth noting the titles that are far less embedded in the network: \newspaper{The Colored American} and \newspaper{The Christian Recorder}, which rarely lead \emph{or} follow. This does not imply that they were not influential in their own right, but it does imply that their discourse was not connected to the discourse of this larger, white-dominated print network. It is quite possible --- as has been suggested in the qualitative scholarship on the subject --- that the conversations documented in these papers stood apart from the dominant, white-led abolitionist discourse. Future research might examine these newspapers for additional evidence of what the conversation about abolition looked like in a predominately Black print sphere.    

\subsection{Network centrality}
\begin{figure}
    \centering
    \includegraphics[width=0.6\textwidth]{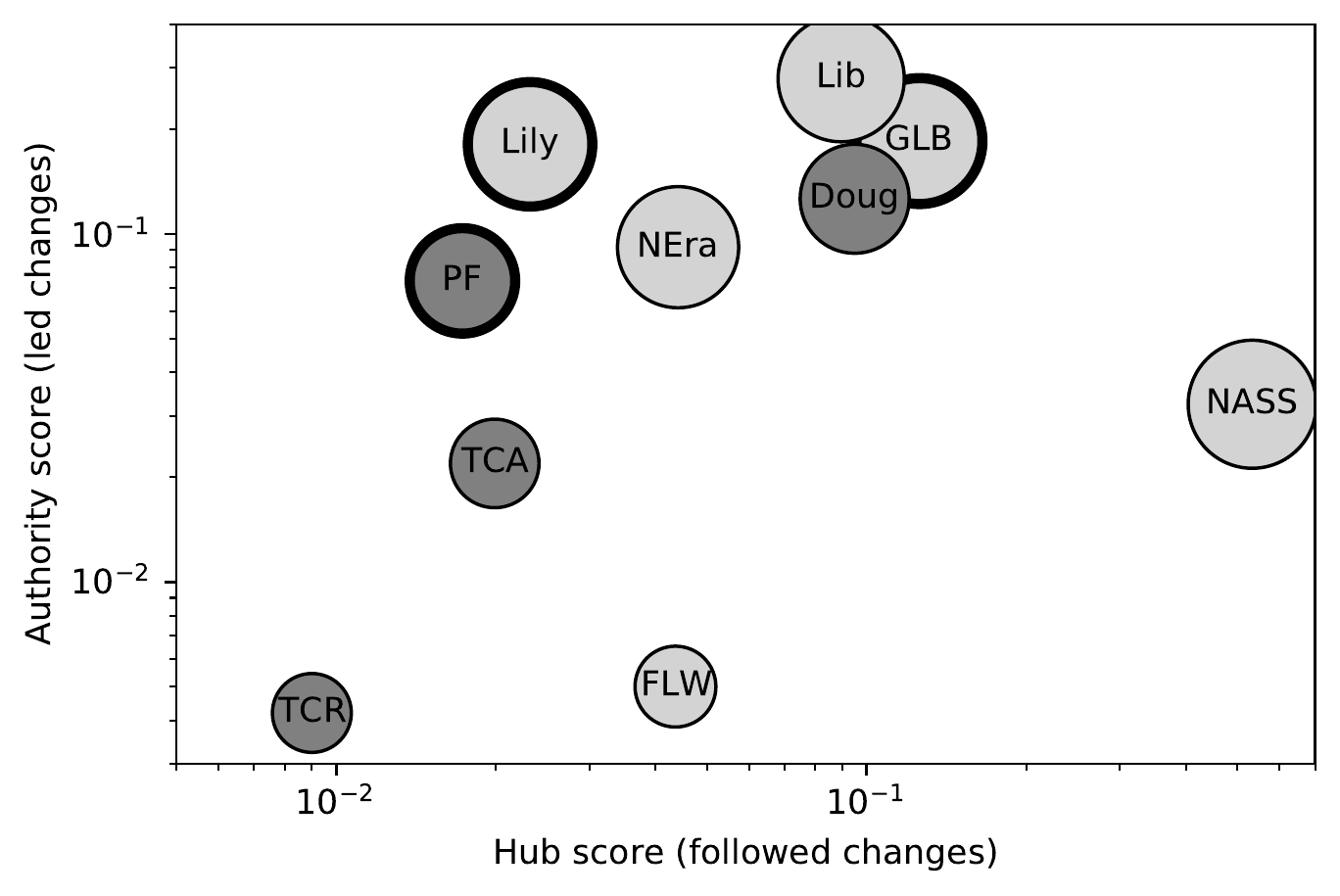}
    \caption{Aggregate network statistics. For each newspaper, the Hub Score roughly corresponds to how often it followed ongoing semantic changes, and the Authority Score to how often it led; node size corresponds to pagerank. Node abbreviations are given in \autoref{fig:spring_layout}. As in \autoref{fig:spring_layout}, newspapers with Black editors are shown in dark gray, and newspapers that were predominantly edited by and/or aimed at women have darker outlines.}
    \label{fig:network_aggregates}
\end{figure}

Finally, to better understand the holistic roles of each newspaper in the network, we now move to the more sophisticated network analytic methods of \emph{pagerank} and \emph{HITS}, described above. We first consider pagerank, which is based on the number of terms on which each newspaper leads, but which especially rewards newspaper who lead other leaders. High pagerank newspapers can therefore be viewed as centralized consolidators of influence. Many of the titles previously discussed exhibit high pagerank scores (shown by node size in \autoref{fig:network_aggregates}), including \newspaper{The National Anti-Slavery Standard} and \newspaper{The Liberator}. Again we see \newspaper{Godey's Lady's Book} and \newspaper{The Lily}, as well as \newspaper{The National Era}, suggesting that these papers each played a central role in consolidating the abolitionist discourse.

The HITS metrics (Hub and Authority scores, shown by position in \autoref{fig:network_aggregates}) help to clarify this relationship by differentiate several distinct groups, as well as some outliers. Once again, \newspaper{National Anti-Slavery Standard} stands out with the largest Hub score by far, indicating its position as a fast follower (but not originator) of semantic change. We also find, conversely, that \newspaper{The Liberator} stands out with the largest Authority score, indicating its position as a leader. A cluster of newspapers --- \newspaper{Godey's Lady's Book}, the \newspaper{Douglass Papers}, as well as the \newspaper{The Liberator} --- scored high on both metrics, indicating their active participation as both leaders and followers of semantic change. 

Returning to our initial focus on leadership, there are two titles with high Authority scores but low Hub scores, indicating that they lead but rarely follow. These are the two titles indicated at the outset of the essay: \newspaper{The Lily} and \newspaper{The Provincial Freeman}, both of which were edited by women. That \newspaper{The Lily}, women's suffrage paper, is so influential as a leader in this network of abolitionist newspapers is a finding that points to the need for additional research, as it might suggest a potentially revisionist narrative to the story that has the white women's suffrage movement departing from --- and, at times, explicitly opposing --- the abolitionist cause. 

That the \newspaper{Provincial Freeman} is the second title on this list is more intriguing still. Upon announcing her departure from the paper, Mary Ann Shadd, the newspaper's founder and editor, lamented that ``Few, if any females had had to contend against the same business'' that she had faced, referring to the criticism she had received as a result of her hard-hitting editorial style \citep{Shadd_1855}. She despairs at the lack of recognition that her paper had commanded, in spite of her having ``broken the Editorial ice'' for ``colored women everywhere'' by serving as the first Black woman to edit a newspaper in all of North America. And yet here in this chart is evidence of the influence that Shadd indeed commanded through her editorial work. Long after \newspaper{The Provincial Freeman} ceased publication, we can return to its contents with fresh eyes, employing quantitative methods to surface the semantic leadership that characterized the paper all along.
\section{Conclusion}
\label{sec:conclusion}

The archive of abolition will always be bound by the historical forces that contributed to its creation --- what Michel-Rolph~\citep{Trouillot_1995} has described as the ``silences'' of the past. But these silences have not stopped scholars from attempting to derive meaning from the documents the archive does contain. On the contrary, the constraints of the archive have prompted a range of scholars to develop powerful new methods for the writing of history, methods which can expand our knowledge of the past. 

With this paper, we aim to contribute to these efforts. We argue that the conceptual undercurrents of the abolitionist movement of the nineteenth-century United States can be characterized in part by a consideration of \emph{neighbors}: as the meanings of words shift over time, these shifts are reflected in statistical changes in the distributions over neighboring terms; similarly, the role of individual newspapers with respect to these changes is characterized by the extent to which they led or followed other newspapers in the adoption of these changes. In some cases, these semantic changes are clearly at the core of the concerns of the abolitionist movement, as in the case of terms like \example{justice} and \example{rights}. Indeed, such examples are so central to the narrative of abolition that they merit close individual analysis through non-computational techniques. Yet the bulk of the semantic changes uncovered by our method are not of this type --- they are ``everyday'' words like \example{busy} and  \example{bold}, \example{ears} and \example{arms}. To understand the story told by these changes, we turn to aggregation, identifying newspapers that consistently led their neighbors in the adoption (and in some cases, instigation) of these shifts. This aggregate view confirms some existing intuitions: for example, \newspaper{The Liberator}, known for its role on the movement's vanguard, is found at the center (\autoref{fig:spring_layout}), while \newspaper{Frank Leslie's Weekly}, which was intentionally less political, is placed at the periphery. Aggregation also enables the use of network statistics to distinguish classes of participants: leaders, like \newspaper{The Lily}; fast followers, like \newspaper{National Anti-Slavery Standard}; and outsiders, like \newspaper{The Christian Recorder}. The aggregate view also proposes some more unexpected results, such as the prominence of the \newspaper{Godey's Lady's Book}, a magazine that sought to position itself apart from the political fray.

From a methodological perspective, a key intervention in this work is the use of randomization to ensure that are findings are robust. This was necessary for two reasons. First, although our dataset is large, we are interested in relatively rare phenomena --- the appearances of individual words in specific linguistic contexts by each newspaper at several different points in time. In any finite dataset, rare phenomena may give rise to spurious correlations which are not represent of meaningful underlying trends. Second, our dataset is heterogeneous: because some newspapers publish earlier and more often than others, temporal methods for detecting influence will inherently favor these newspapers. Randomization addresses both of these issues by comparing our quantitative results against the distribution of results obtained from a large set of alternative datasets in which no semantic leadership is possible by construction. This methodology of control by randomization is broadly applicable, and is especially applicable to analysis of temporal phenomena, which is difficult to validate using traditional hypothesis testing~\citep[e.g.,][]{dubossarsky2017outta,dubossarsky2019time}.

Our hope is that the results presented here --- the specific semantic changes, the newspapers that exhibit semantic leadership (or that follow those leaders), and the overall picture of the network of semantic influence among these various papers --- will become additional evidence that scholars working on the archive of abolition can incorporate into their future work. Our findings about the lexical semantic leadership of two newspapers edited by women, one white and one Black, certainly seem to prompt further investigation. The network of lexical semantic leadership among the members of the Black press also seems worthy of additional inquiry. And while any model developed with a particular archive in mind must be translated to other contexts with intention and care, we believe that our model, as documented in this essay, can be applied to other archives in which questions of influence run up against the constraints of the historical record, potentially enabling new narratives about the actors and agents involved in advancing a shared social, cultural, or political goal.
\section{Acknowledgments}
\label{sec:acknowledgments}
This paper was published in the Journal of Cultural Analytics~\citep{soni2021abolitionist}. The authors thank Andrew Piper, the editor of the journal, David Smith and Richard Jean So, who reviewed our paper and provided valuable feedback to improve the paper. The initial corpus creation effort was performed as part of the Interactive Topic Model and Metadata Visualization (TOME) project, supported by NEH Office of Digital Humanities Grant HD-51705-13. This research was additionally supported by NSF IIS-1452443.

\bibliographystyle{plainnat}
\bibliography{references}  

\end{document}